\documentclass{article}

\usepackage{arxiv}

\usepackage{enumitem}
\usepackage{graphicx}
\usepackage{amsmath}
\usepackage{multirow}
\usepackage{placeins}  

\usepackage[utf8]{inputenc} 
\usepackage[T1]{fontenc}    

\usepackage{hyperref}       
\hypersetup{
    pdfborder={0 0 1},
    citebordercolor={0 1 0}
}
\usepackage[capitalize]{cleveref}

\usepackage{url}            
\usepackage{booktabs}       
\usepackage{amsfonts}       
\usepackage{nicefrac}       
\usepackage{microtype}      
\usepackage{xcolor}         

\makeatletter
\newif\iffirstsection
\firstsectiontrue
\let\origsection\section
\renewcommand{\section}{%
    \iffirstsection
        \global\firstsectionfalse
    \else
        \FloatBarrier
    \fi
    \origsection
}
\makeatother

\title{FreeOcc: Decoupling Ego-Motion for Efficient 4D Occupancy Forecasting via Continuous Flow Matching}

\author{%
  Zeping Zhang \\
  University of Ottawa \\
  \texttt{Zeping.Zhang@uottawa.ca} \\
  \And
  Zhuoya Zhao \\
  University of Ottawa \\
  \texttt{zoezhao0427@gmail.com} \\
  \AND
  Samy Metari \\
  National Research Council Canada \\
  \texttt{Samy.Metari@nrc-cnrc.gc.ca} \\
  \And
  Robert Lagani\`{e}re\thanks{Corresponding author.} \\
  University of Ottawa \\
  \texttt{laganier@uottawa.ca} \\
}

\begin{document}

\maketitle
{\renewcommand{\thefootnote}{}\footnotetext{Accepted at the 40th Conference on Neural Information Processing Systems (NeurIPS 2026).}}

\begin{abstract}
Current 4D occupancy world models use heavy cross-attention or deep recurrence to disentangle ego-motion from scene dynamics. This requires significant computational capacity. However, this effort is largely redundant since the ego-vehicle's intended future motion is provided by the downstream motion planner. We present FreeOcc, a lightweight continuous-time world model that introduces explicit $SE(2)$ geometric decoupling into a conditional flow-matching framework. By mathematically factoring out self-motion, we create a purely dynamic feature space. This space supports attention-free temporal fusion and allows physics-constrained ODEs to operate strictly on residual agent kinematics. To maximize deployment efficiency, FreeOcc introduces an uncertainty-aware blending mechanism. This routes well-observed voxels to a single-pass decoder, reserving the ODE solver exclusively for ambiguous, high-uncertainty regions. Requiring only 28.3M parameters—roughly 40\% of comparable models—our optimal ensemble mode (FreeOcc-E) runs at 27 FPS. On nuScenes, FreeOcc achieves state-of-the-art forecasting performance with a 0.47 m Near Field Chamfer Distance (NFCD) and 32.1\% mIoU. Furthermore, the model transfers zero-shot to SemanticKITTI. It also proves highly robust, retaining 92.8\% of its ideal mIoU under significant sensor packet loss and timestamp jitter.
\end{abstract}

\vspace{-0.8em}

\section{Introduction}

Anticipating the future 3D structure of a driving scene is critical for pre-emptive motion planning. Dense 4D occupancy forecasting addresses this by predicting a sequence of future voxelised binary volumes $\{O_{\tau}\}_{\tau=1}^{N}$ directly from recent sensor history, elegantly capturing complex scene geometry without relying on explicit object detection or tracking.

Recent works tackle this via: autoregressive token generation~\cite{zheng2025occworld,zhang2023learning}, continuous flow formulations~\cite{shi2024streamingflow}, geometric proxy representations~\cite{yu2026superocc}, and unsupervised implicit fields~\cite{agro2024uno,yang2023vidar}. However, these methods generally rely on cross-attention or deep recurrence to separate the ego-vehicle's motion from observed scene dynamics. This entanglement inflates parameter counts. It also requires expensive temporal attention and complicates error control at long horizons. Crucially, the ego-vehicle's future trajectory is typically available at inference time as the intended route provided by the motion planner. While explicit $SE(2)$ warping in BEV space is well established~\cite{jaderberg2015spatial}, recent generative world models have largely favoured learned temporal attention instead. In this setting, learned alignment is unnecessary: a single geometric warp factors out ego-motion and leaves the model to focus on residual dynamics, occlusion evolution, and far-field extrapolation. Driven by this insight, we propose \textbf{FreeOcc}. By mathematically factoring out self-motion, our architecture simplifies the forecasting pipeline, yielding three primary contributions:

\begin{itemize}[nosep,leftmargin=1.5em]
\item \textbf{Attention-Free Temporal Fusion:} Removing ego-motion allows a lightweight ConvGRU to effectively capture residual dynamics. By avoiding dense cross-attention, FreeOcc processes a $200 \times 200 \times 16$ voxel resolution at 27 FPS. Furthermore, it achieves this with only 28.3M parameters, which is roughly 40\% of the size of comparable world models.
\item \textbf{Physics-Constrained Generation:} The motion-decoupled feature space stabilizes the conditioning for flow matching. Physical constraints—such as velocity smoothness, mass conservation, and ray-traced free-space—can be applied directly to agent kinematics without relying on restrictive geometric proxies.
\item \textbf{Uncertainty-Aware Routing:} Decoupling self-motion separates true scene ambiguity—such as dynamic objects and occlusion boundaries—from misalignment artefacts. The decoder’s per-voxel confidence is used to route well-observed regions to a single-pass decoder, while the continuous ODE solver handles highly uncertain areas.
\end{itemize}

On nuScenes, FreeOcc achieves state-of-the-art forecasting performance (0.47 m NFCD, 32.1\% mIoU) and transfers zero-shot to SemanticKITTI. The model remains robust under significant sensor dropout and temporal jitter, retaining 92.8\% of its ideal mIoU.

\section{Related Work}

\paragraph{4D occupancy forecasting.} 
Recent world models approach future scene prediction through autoregressive token generation~\cite{zheng2025occworld}, discrete diffusion~\cite{zhang2023learning}, implicit fields~\cite{agro2024uno}, or camera-based point cloud forecasting~\cite{yang2023vidar}, and have recently been scaled to long-horizon occupancy rollouts for multi-kilometer simulation~\cite{liu2026occsim}. However, these paradigms typically dedicate massive capacity to entangling ego-motion and scene dynamics via deep recurrence or cross-attention. FreeOcc bypasses this bottleneck by explicitly factoring out self-motion via algebraic $SE(2)$ warping (see Section 3.4), allowing it to directly predict dense voxel grids via physics-constrained continuous flow matching (CFM) without relying on discrete tokenisation or implicit rendering.

\paragraph{Flow matching and continuous generative models.} 
Continuous normalising flows trained via flow matching~\cite{lipman2023flow} offer a simulation-free generative objective. Recent works like Streaming Flow~\cite{shi2024streamingflow}, and latent flow models~\cite{liu2025towards} adapt this to occupancy and LiDAR streams. Yet, they still rely on computationally heavy dense cross-attention for temporal alignment. FreeOcc preserves the continuous expressivity of flow matching but replaces the attention bottleneck with an explicit geometric warp, further regularising the generative trajectory with physics-informed constraints (velocity smoothness, mass conservation, ray-traced free-space).

\paragraph{Physics-informed perception.} 
While integrating physical priors into perception is well-established—such as range-aware encoding~\cite{lang2019pointpillars} or heteroscedastic uncertainty~\cite{kendall2017uncertainties}—applying structural constraints to dense 4D forecasting remains challenging. Recent concurrent works like SuperOcc~\cite{yu2026superocc} and DTT~\cite{xu2025delta} enforce geometric structure by restricting the output space to predefined proxies (e.g., superquadrics or delta residuals), which inevitably sacrifices fine-grained fidelity on irregular shapes. Rather than simplifying the output representation, FreeOcc guides the full-resolution generative trajectory at the \emph{loss} level. This ensures physically plausible dynamics while retaining exact voxel-level structural fidelity.


\section{Methodology}
\label{sec:methodology}
FreeOcc forecasts dense 4D occupancy from LiDAR with an explicit task division: geometry handles ego-motion, while learning handles residual scene dynamics. By mathematically isolating the vehicle's known trajectory via differentiable $\mathrm{SE}(2)$ warping, we remove the need for learned temporal alignment, allowing the generative model to focus entirely on independently moving agents, occlusion evolution, and long-horizon uncertainty.


\subsection{Overall Architecture}
\label{subsec:architecture}

\begin{figure*}[t]
    \centering
    \includegraphics[width=1.0\textwidth, trim=10mm 90mm 10mm 87mm, clip]{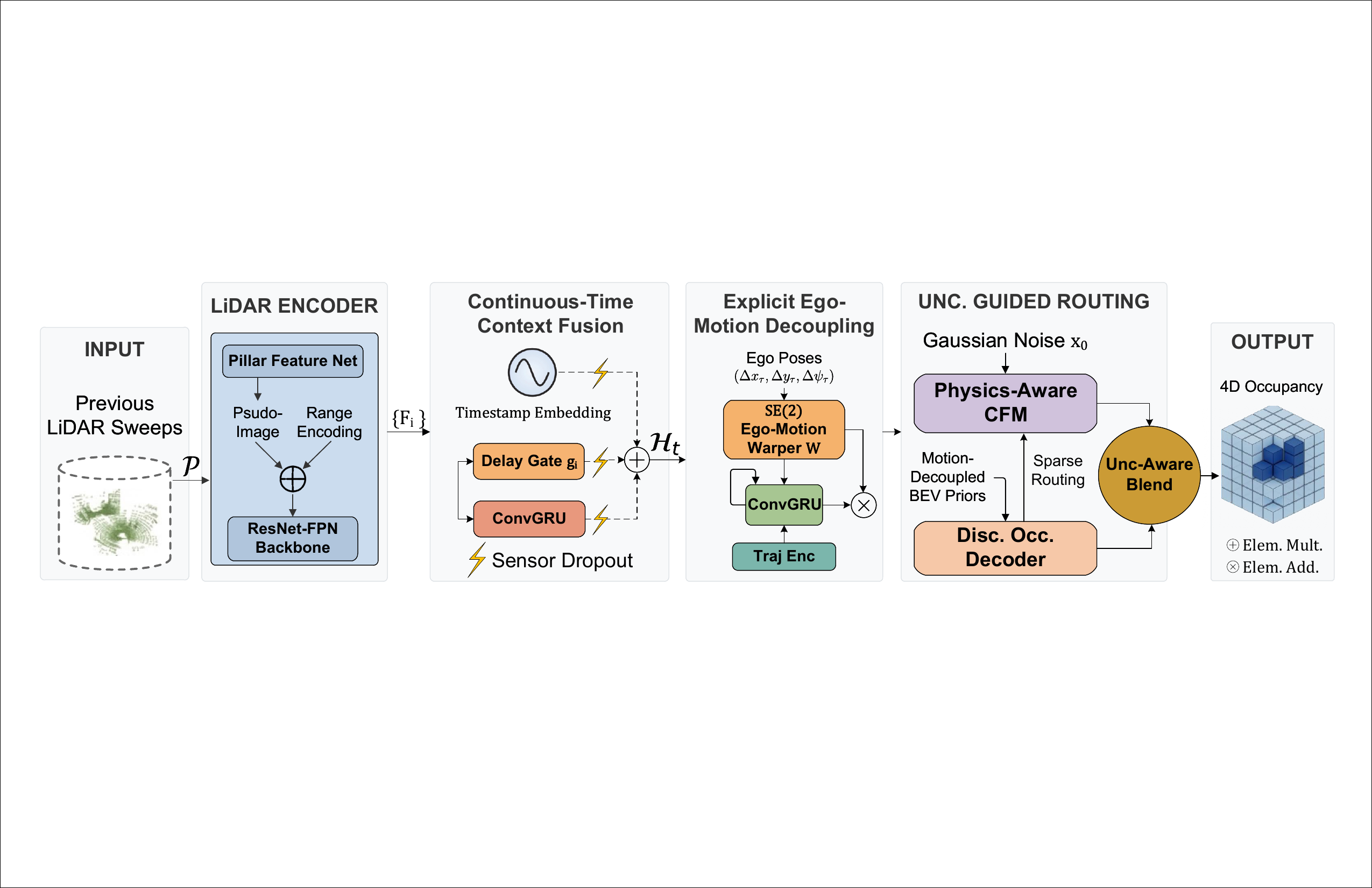}
    \caption{Overview of FreeOcc. Previous sweeps are encoded into BEV, fused by a continuous-time temporal aggregator, propagated to future ego frames by $\mathrm{SE}(2)$ warping, and decoded by the Discriminative Occupancy Decoder plus the Physics-Aware Continuous Flow Matching. The final prediction blends both paths using decoder uncertainty.}
    \label{fig:main_arch}
\end{figure*}

The architecture of FreeOcc (depicted in \Cref{fig:main_arch}) is driven by this principle of explicit geometric decoupling, operating in three distinct phases:

\begin{itemize}[nosep,leftmargin=1.5em]
    \item \textbf{Continuous-Time Context Fusion (\cref{subsec:encoder,subsec:aggregator}):} Given $K$ asynchronous previous sweeps $\mathcal{P}=\{P_{t-K+1},\dots,P_{t}\}$, we encode each sweep into BEV and fuse them via timestamp-aware recurrent aggregation into a unified present-state conditioning map $\mathcal{H}_t\in\mathbb{R}^{C\times H\times W}$.
    
    \item \textbf{Explicit Ego-Motion Decoupling (\cref{subsec:propagator}):} Rather than learning temporal alignment, an ego-aware propagator explicitly warps $\mathcal{H}_t$ to future coordinate frames. The resulting BEV priors $\{C_\tau\}_{\tau=1}^{N}$ isolate the residual scene changes.
    
    \item \textbf{Uncertainty-Guided Dual-Path Generation (\cref{subsec:flow_matching,subsec:decoder,subsec:blending}):} Each prior $C_\tau$ is decoded through two branches to maintain inference speed. Confident, well-observed regions with lower uncertainty pass through a fast discriminative decoder. The resulting uncertainty map then selectively routes only the highly uncertain areas to a physics-aware conditional flow-matching ODE.
\end{itemize}

We use the following notation throughout. BEV features have spatial size $H\!\times\!W$ and channel width $C$; voxel grids have size $X\!\times\!Y\!\times\!Z$ at resolution $\delta$ (m/voxel). The current time is $t$, the future index is $\tau\in\{1,\dots,N\}$, and the internal flow time is $s\in[0,1]$.

\subsection{LiDAR Encoder with Sensor-Range Encoding}
\label{subsec:encoder}

Each sweep $P_i=\{p_j\}_{j=1}^{M_i}$, where $M_i$ is the number of points in the sweep and $p_j\in\mathbb{R}^5$ (coordinates, intensity, ring), is first processed by a Pillar Feature Net~\cite{lang2019pointpillars} (see \Cref{fig:main_arch}), which encodes points into a sparse pillar representation. Specifically, a shared three-layer MLP maps each point within a pillar to $\mathbb{R}^C$, and per-pillar max-pooling yields a pseudo-image $\bar{F}_i \in \mathbb{R}^{C \times H \times W}$.

We further add a learned sinusoidal range encoding $\rho(r_{hw})$ to each BEV cell, giving the encoder an explicit distance prior that reflects the drop in LiDAR sampling density with range. A lightweight ResNet--FPN backbone~\cite{he2016deep,lin2017feature} then produces the per-sweep feature $F_i\in\mathbb{R}^{C\times H\times W}$; the exact encoding formula is given in ~\cref{sec:appx_range_encoding}.

\subsection{Continuous-Time Temporal Aggregator}
\label{subsec:aggregator}

After encoding the individual sweeps into spatial features $F_i$, the model must aggregate them over time. Real-world LiDAR data suffers from scheduling jitter, delays and packet loss. Instead of learning temporal alignment with attention, FreeOcc encodes each sweep at its measured arrival time and lets a lightweight ConvGRU absorb the irregular history.

For the $i$-th sweep at absolute time $t_i$, we compute $\Delta t_i = t_i - t \leq 0$ and encode it with an exponentially spaced sinusoidal timestamp embedding $\phi(\Delta t_i)$ (detailed in ~\cref{sec:appx_method_details}). We then refine it with a learned projection $\psi:\mathbb{R}^C\!\to\!\mathbb{R}^C$. A delay-aware confidence gate discounts stale observations:


\begin{equation}
    \label{eq:delay_gate}
    g_i = \sigma\!\bigl(\mathrm{ReLU}(w_1\;\psi(\phi(\Delta t_i))) + b\bigr) \;\in\; (0,1),
\end{equation}
where $\sigma$ denotes the sigmoid activation function, $w_1$ is a learned weight matrix, and $b$ is an additive bias initialized to $+2$ ($\sigma(2)\approx 0.88$).

The time-aware feature is $\tilde{F}_i = g_i \cdot F_i + \psi(\phi(\Delta t_i))$.
During training, we additionally perturb sweep timestamps and randomly drop non-current sweeps so the GRU sees both asynchronous and incomplete histories. This combination improves robustness to timing noise and sensor dropout without adding attention parameters; the exact augmentation settings are given in ~\cref{sec:appx_augmentation}. The sequence $\{\tilde{F}_i\}$ is fused by a two-layer ConvGRU~\cite{cho2014learning}, and the final hidden state is layer-normalised to yield $\mathcal{H}_t\in\mathbb{R}^{C\times H\times W}$.

\subsection{Ego-Aware Future Propagator: Geometric Decoupling}
\label{subsec:propagator}

\begin{figure}[t]
    \centering
    \includegraphics[width=0.9\textwidth, trim=63mm 69mm 80mm 94mm, clip]{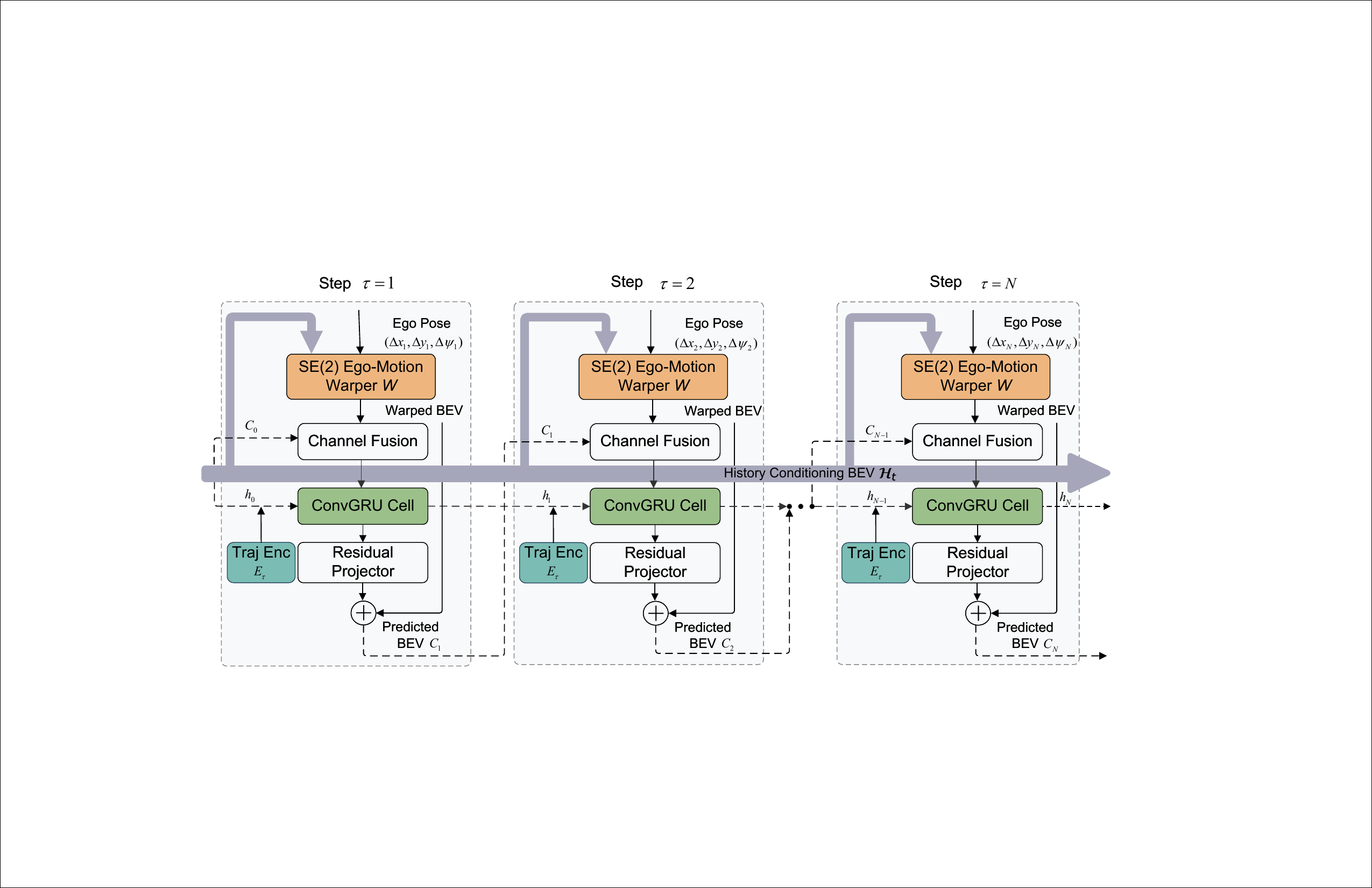}
    \caption{Ego-aware future propagator. At each horizon, the warped BEV—obtained via SE(2) ego-motion alignment—is explicitly added to the residual scene dynamics predicted by a ConvGRU and a residual projector.}
    \label{fig:propagator}
\end{figure}

With the historical context fused into the present-state representation $\mathcal{H}_t$, FreeOcc uses explicit geometric warping (\Cref{fig:propagator}) to remove the dominant ego-induced motion before prediction. The resulting feature space is easier to learn for both the ConvGRU and the velocity network $v_\theta$: the static structure is aligned by construction, and the remaining updates correspond to residual scene dynamics.

Given the intended future relative ego-pose $(\Delta x_\tau, \Delta y_\tau, \Delta\psi_\tau)$---recorded from odometry during training and provided by the onboard motion planner at deployment---we construct a $2\!\times\!3$ affine matrix:
\begin{equation}
    \label{eq:affine}
    A_\tau = \begin{bmatrix}
        \cos\Delta\psi_\tau & \sin\Delta\psi_\tau & \Delta y_\tau / R \\
        -\sin\Delta\psi_\tau & \cos\Delta\psi_\tau & \Delta x_\tau / R
    \end{bmatrix},
\end{equation}
where $R$ is the BEV half-range.
The warping $\mathcal{W}$ applies this via bilinear grid sampling~\cite{jaderberg2015spatial}, mapping $\mathcal{H}_t$ to the future coordinate frame.

At each step $\tau$, a ConvGRU cell fuses three inputs (the warped BEV, the previous output $C_{\tau-1}$, and a trajectory encoding $E_\tau = \mathrm{MLP}(\Delta x_\tau, \Delta y_\tau, \Delta\psi_\tau)\in\mathbb{R}^C$) via channel fusion ($1 \times 1$ convolution):
\begin{equation}
    \label{eq:convgru}
    h_\tau = \mathrm{ConvGRU}\!\bigl(\mathrm{Conv}_{1\times 1}[\mathcal{W}(\mathcal{H}_t, A_\tau) \,\Vert\, C_{\tau-1} \,\Vert\, E_\tau],\; h_{\tau-1}\bigr),
\end{equation}
with $h_0 = C_0 = \mathcal{H}_t$.
A residual skip from the warped BEV ensures the ConvGRU learns only the \emph{residual deviation}:
\begin{equation}
    \label{eq:residual}
    C_\tau = \mathrm{Proj}(h_\tau) + \mathcal{W}(\mathcal{H}_t, A_\tau),
\end{equation}
where Proj denotes the Residual Projector (a $3 \times 3$ convolution followed by GroupNorm and ReLU), as illustrated in \Cref{fig:propagator}.
This factorisation separates independent-agent dynamics from ego-induced geometric shift and makes the velocity-field constraints in \cref{subsec:flow_matching} well-defined.

\subsection{Physics-Aware Continuous Flow Matching}
\label{subsec:flow_matching}

Because the propagator has factored out the ego-vehicle's motion, the resulting priors $C_\tau$ isolate the independent agent dynamics. This decoupled feature space provides a highly stable condition for the optimal-transport flow matching problem~\cite{lipman2023flow}.
Let $x_1\in\{0,1\}^{X\times Y\times Z}$ denote the ground-truth binary occupancy and $x_0\sim\mathcal{N}(0,I)$.
The conditional probability path is:
\begin{equation}
    x_s = (1-s)\,x_0 + s\,x_1, \quad s\in[0,1],
\end{equation}
with target velocity $u(s)=x_1-x_0$.
A 3D U-Net velocity network $v_\theta$ is trained to predict this velocity. Let $\tilde{C}_\tau$ be the height-expanded version of the propagated BEV prior $C_\tau$; the velocity network is directly conditioned on $\tilde{C}_\tau$.

Because the propagator already removes ego motion, the velocity field $v_\theta$ operates on residual agent dynamics. We regularise it with an anisotropic total variation so nearby voxels on the same rigid body move coherently:
\begin{equation}
    \label{eq:vel_smooth}
    \mathcal{L}_\mathrm{vel} = \frac{1}{3}\sum_{d\in\{x,y,z\}} \frac{1}{|\Omega_d|}\sum\bigl|v_\theta^{(d+1)} - v_\theta^{(d)}\bigr|,
\end{equation}
where the sum runs over adjacent voxels along each axis.

\paragraph{Cooperative uncertainty-guided training.}
We couple the decoder's uncertainty to the generative branch so ODE capacity is spent where single-pass prediction is least reliable. Let $\hat{s}_v$ denote the detached per-voxel log-variance predicted by the decoder (\cref{subsec:decoder}). The flow-matching loss is weighted as
\begin{equation}
    \label{eq:cfm_loss}
    \mathcal{L}_\mathrm{FM} = \mathbb{E}_{s,x_0}\!\bigl[\,w(x_1, \hat{s}_v)\,\bigl\|v_\theta(x_s, s, \tilde{C}_\tau) - u(s)\bigr\|_2^2\,\bigr],
\end{equation}
where
\begin{equation}
    \tilde{u}_v = \operatorname{clamp}\!\left(\frac{\hat{s}_v-\mu_{\mathrm{low}}}{\mu_{\mathrm{high}}-\mu_{\mathrm{low}}},\,0,\,1\right),
    \qquad
    w(x_1, \hat{s}_v) = (1+x_1)\cdot(1+\tilde{u}_v).
\end{equation}
Here $\mu_{\mathrm{low}}$ and $\mu_{\mathrm{high}}$ are EMA estimates of the 5th and 95th percentiles of $\hat{s}_v$ over training batches. The first factor prevents the trivial all-empty solution---most voxels are air or free space, so an unweighted model can reduce loss by predicting empty almost everywhere. The second factor upweights voxels where the decoder is uncertain, such as boundaries, dynamic objects, and far-field regions. Network details are given in ~\cref{sec:appx_flow_network}.

At inference, the ODE is solved from $s=0$ to $s=1$:
\begin{equation}
    \frac{dx}{ds} = v_\theta(x, s, \tilde{C}_\tau), \quad x(0)\sim\mathcal{N}(0,I),
\end{equation}
using a fixed-step midpoint integrator with $S$ steps (default $S=10$), yielding $\hat{O}^\mathrm{ode}_\tau = \mathrm{clamp}(x(1), [0,1])$. While continuous flow matching provides exceptional fidelity for complex scene dynamics, solving the ODE across the entire volume is computationally redundant for static, well-observed geometry. Therefore, we couple our flow-matching objective with a fast discriminative decoder (\cref{subsec:decoder}), directing the ODE's capacity strictly toward regions of high uncertainty.
\subsection{Discriminative Occupancy Decoder}
\label{subsec:decoder}

The second path lifts each BEV prior $C_\tau$ to a voxel volume with a shallow BEV-to-3D decoder and predicts both occupancy logits $\hat{y}_\tau\in\mathbb{R}^{X\times Y\times Z}$ and per-voxel log-variance $\hat{s}_\tau\in\mathbb{R}^{X\times Y\times Z}$. We add a future-step embedding before lifting; the exact decoder and head configuration is given in ~\cref{sec:appx_decoder}.
The decoder prediction is $\hat{O}^\mathrm{dec}_\tau = \sigma(\hat{y}_\tau)$.

The per-voxel log-variance $\hat{s}_\tau$ serves a dual role: it enters the cooperative flow matching loss (\cref{subsec:flow_matching}) during training, and drives uncertainty-aware blending at inference (\cref{subsec:blending}).

\subsection{Uncertainty-Aware Ensemble Blending}
\label{subsec:blending}

The two generation paths offer complementary strengths: the decoder provides fast, high-recall predictions in well-observed regions, while the ODE yields high-precision samples at the cost of iterative computation, excelling in ambiguous or dynamic areas.
Rather than fixed blending (max or mean), we propose uncertainty-aware blending, which uses the decoder's own per-voxel aleatoric uncertainty to route each voxel to the appropriate path.

We convert the predicted log-variance $\hat{s}_v$ into a bounded confidence weight:
\begin{equation}
    \label{eq:conf_weight}
    w_v = \sigma(-\hat{s}_v) \;\in\; (0, 1).
\end{equation}
The final occupancy probability is a confidence-weighted interpolation:
\begin{equation}
    \label{eq:unc_blend}
    \hat{O}_\tau(v) = w_v\,\hat{O}^\mathrm{dec}_\tau(v) + (1 - w_v)\,\hat{O}^\mathrm{ode}_\tau(v).
\end{equation}
Low log-variance implies high confidence and therefore high $w_v$, so the blend favours the fast decoder in well-observed regions. Higher log-variance lowers $w_v$ and shifts weight toward the ODE in ambiguous or far-field areas.
The uncertainty head is trained jointly with the decoder, so the routing signal improves alongside prediction quality.

FreeOcc supports three inference modes with distinct speed/accuracy trade-offs:
\begin{itemize}[nosep,leftmargin=1.5em]
    \item \textbf{FreeOcc-D} (Decoder-only): single forward pass, real-time;
    \item \textbf{FreeOcc-G} (Generative ODE-only): $S$-step ODE, highest precision;
    \item \textbf{FreeOcc-E} (Ensemble): uncertainty-aware blending, best IoU at acceptable latency;
\end{itemize}

\subsection{Physics-Aware Training Objectives}
\label{subsec:losses}

FreeOcc is trained end-to-end with a composite loss coupling the flow-matching objective ($\mathcal{L}_\mathrm{FM}$) with three specialized regularization groups:
\begin{equation}
    \label{eq:total_loss}
    \begin{split}
        \mathcal{L} = \mathcal{L}_\mathrm{FM} 
        &+ \underbrace{\lambda_\mathrm{occ}\mathcal{L}_\mathrm{occ} + \lambda_\mathrm{lov}\mathcal{L}_\mathrm{Lov} + \lambda_\mathrm{dice}\mathcal{L}_\mathrm{Dice} + \lambda_\mathrm{sp}\mathcal{L}_\mathrm{sp} + \lambda_\mathrm{tc}\mathcal{L}_\mathrm{tc}}_{\text{Structural Fidelity}} \\
        &+ \underbrace{\lambda_\mathrm{ray}\mathcal{L}_\mathrm{ray} + \lambda_\mathrm{cons}\mathcal{L}_\mathrm{cons} + \lambda_\mathrm{vel}\mathcal{L}_\mathrm{vel} + \lambda_\mathrm{rw}\mathcal{L}_\mathrm{rw}}_{\text{Physical Priors}} 
        + \underbrace{\lambda_\mathrm{unc}\mathcal{L}_\mathrm{unc}}_{\text{Uncertainty}},
    \end{split}
\end{equation}
where $\lambda$ denotes the hyperparameter weights. Structural fidelity supervises the discriminative decoder, while physical priors enforce ray-traced free-space, mass conservation, and piecewise-smooth kinematics (\cref{eq:vel_smooth}). For uncertainty calibration, we employ a heteroscedastic classification loss:
\begin{equation}
    \mathcal{L}_\mathrm{unc} = \frac{1}{|\Omega|}\sum_{ijk}\left[\exp(-\hat{s}_{ijk})\,\ell_{\text{BCE}}(\hat{y}_{ijk}, y_{ijk}) + \beta\hat{s}_{ijk}\right],
\end{equation}
where $\ell_{\text{BCE}}$ denotes the binary cross-entropy loss applied to the predicted occupancy logits $\hat{y}_{ijk}$, and $\beta=0.5$ is a fixed regularization penalty to prevent degenerate solutions. Detailed descriptions for each term are provided in ~\cref{sec:appx_loss_defs}. We omit Occ3D~\cite{tian2024occ3d} because its focus on completing unobservable hidden geometry misaligns with our target task of sensor-grounded forecasting (see \cref{sec:appx_occ3d} for a detailed discussion).

\section{Experiments}
\label{sec:experiments}

\subsection{Experimental Setup}
\label{subsec:setup}

\paragraph{Datasets.}
We train on nuScenes~\cite{caesar2020nuscenes} and evaluate both in-domain and zero-shot on SemanticKITTI~\cite{behley2019semantickitti}. For the cross-domain setting, we apply simple geometric calibration to the SemanticKITTI point clouds before evaluation. Dataset splits and calibration details are given in ~\cref{sec:appx_metrics}.

\paragraph{Voxel Specification.} We use a $[-40, 40]\text{m}$ spatial range ($200\!\times\!200\!\times\!16$ grid at $\delta\!=\!0.4$\,m resolution) with $N\!=\!6$ future and $K\!=\!4$ history steps, but explicitly train a dedicated $[-70, 70]\text{m}$ variant to ensure fair spatial comparisons against point-based baselines in Table~\ref{tab:geometry_forecasting}.

\paragraph{Metrics \& Evaluation Protocol.}
\label{para:metrics}
We report two complementary metric tracks:
\begin{itemize}[nosep,leftmargin=1.5em]
    \item Voxel metrics: class-agnostic mIoU, precision, recall, and F1 on binarized occupancy grids;
    \item Point metrics: NFCD, L1 depth error, and AbsRel following prior point-forecasting work~\cite{agro2024uno,zhang2023learning}.
\end{itemize}
For point metrics, we convert predicted occupied voxels to voxel-centre point clouds. Metric alignment details are given in ~\cref{sec:appx_metrics}.

\paragraph{Implementation.}
FreeOcc (28.3M parameters) is trained end-to-end on 4 NVIDIA RTX 4090 GPUs for 10 epochs with AdamW ($\mathrm{lr}\!=\!3\!\times\!10^{-4}$, cosine decay, 500-step warmup, wd\,=\,0.01), batch size 2 with $2\times$ gradient accumulation, and FP16 mixed precision.
ODE inference: $S\!=\!10$ midpoint steps.

\paragraph{Baselines.}
We compare against representative voxel- and point-based methods:
\begin{itemize}[nosep,leftmargin=1.5em]
    \item 4D-Occ~\cite{khurana2023point}, OccWorld~\cite{zheng2025occworld}, and ViDAR~\cite{yang2023vidar} for dense occupancy comparison;
    \item Copilot4D~\cite{zhang2023learning} and UnO~\cite{agro2024uno} for point-based structural geometry;
    \item \textbf{Static Copy}: Current-frame GT ego-warped to each future step, serving as a no-learned-dynamics lower bound.
\end{itemize}
For cross-modality fairness, point outputs are voxelized when evaluated on occupancy metrics, and our voxel predictions are converted to point clouds when evaluated on NFCD/L1. OccWorld is trained from scratch using the released implementation. Copilot4D and UnO are compared only on point metrics using values reported in their original papers$^\dagger$.

\paragraph{Note on recent concurrent work.}
Concurrent works SuperOcc~\cite{yu2026superocc} and DTT~\cite{xu2025delta} rely on predefined geometric proxies. Crucially, SuperOcc targets current-frame 3D prediction rather than 4D forecasting, and DTT currently lacks publicly released code. Therefore, neither is directly applicable as a forecasting baseline under our protocol, and we include them strictly as conceptual comparisons.

\subsection{Main Results on nuScenes}
\label{subsec:main_results}

We report point-based geometry forecasting and voxel-based occupancy forecasting separately.

\begin{table}[t]
    \centering
    \caption{3D Geometry Forecasting on nuScenes. Evaluated via Near Field Chamfer Distance (NFCD, m) $\downarrow$, L1 Depth Error (m) $\downarrow$, and Absolute Relative Error (AbsRel, \%) $\downarrow$.}
    \label{tab:geometry_forecasting}
    \small
    \begin{tabular}{l c c c | c c c}
    \toprule
    \multirow{2}{*}{Method} & \multicolumn{3}{c|}{1.0s ($\tau{=}2$)} & \multicolumn{3}{c}{3.0s ($\tau{=}6$)} \\
    \cmidrule{2-7}
    & NFCD & L1 & AbsRel & NFCD & L1 & AbsRel \\
    \midrule
    4D-Occ$^\dagger$~\cite{khurana2023point}        & 1.41 & 1.40 & 10.37 & 1.40 & 1.71 & 13.48 \\
    Copilot4D$^\dagger$~\cite{zhang2023learning} & \textbf{0.36} & 1.30 & 8.58  & 0.58 & 1.51 & 10.38 \\
    UnO$^\dagger$~\cite{agro2024uno}           & --   & --   & --    & 0.89 & 1.15 & 8.50  \\
    ViDAR$^\dagger$~\cite{yang2023vidar}          & 0.85 & 1.55 & 11.50 & 1.18 & 1.82 & 14.20 \\
    FreeOcc-D (ours)                              & 0.48 & 1.25 & 8.15  & 0.63 & 1.35 & 9.20  \\
    FreeOcc-G (ours)                              & 0.45 & 1.20 & 7.85  & 0.52 & 1.15 & 8.10  \\
    FreeOcc-E (ours)                              & 0.41 & \textbf{1.18} & \textbf{7.60} & \textbf{0.47} & \textbf{1.08} & \textbf{7.45} \\
    \bottomrule
    \end{tabular}
    
    \vspace{3pt}
    \raggedright \footnotesize
    $^\dagger$Numbers from original papers. L1 Depth Error is an evaluation-only metric computed by projecting predicted occupied voxels to range images and comparing against GT LiDAR depths; no L1 loss is required during training.
\end{table}

\begin{table}[t]
    \centering
    \caption{4D Occupancy Forecasting on nuScenes. Evaluated via Class-Agnostic 3D Voxel mIoU (\%) $\uparrow$.}
    \label{tab:occupancy_forecasting}
    \footnotesize
    \setlength{\tabcolsep}{3.5pt}
    \begin{tabular}{l cccccc | cccc}
    \toprule
    \multirow{2}{*}{Method} & \multicolumn{6}{c|}{Forecasting Horizon mIoU (\%)} & \multicolumn{4}{c}{Average Metrics (\%)} \\
    \cmidrule{2-11}
     & $\tau{=}1$ & $\tau{=}2$ & $\tau{=}3$ & $\tau{=}4$ & $\tau{=}5$ & $\tau{=}6$ & mIoU & Prec. & Rec. & F1 \\
    \midrule
    Static Copy                                   & 31.5 & 22.1 & 14.8 & 10.5 & 8.8 & 7.5 & 15.9 & 34.0 & 23.5 & 27.8 \\
    4D-Occ$^*$~\cite{khurana2023point}            & 35.8 & 27.5 & 22.4 & 19.0 & 16.5 & 14.2 & 22.6 & 39.1 & 31.8 & 35.1 \\
    OccWorld~\cite{zheng2025occworld}             & 39.7 & 34.2 & 27.9 & 26.4 & 23.6 & 21.6 & 28.9 & 50.5 & 40.2 & 44.8 \\
    ViDAR$^*$~\cite{yang2023vidar}                & 30.5 & 24.2 & 20.5 & 17.8 & 15.9 & 14.3 & 20.5 & 39.5 & 29.8 & 34.1 \\
    FreeOcc-D (ours)                              & 40.8 & 34.3 & 29.1 & 25.1 & 21.8 & 19.2 & 28.4 & 41.3 & \textbf{49.4} & 44.9 \\
    FreeOcc-G (ours)                              & 41.9 & 35.2 & 30.2 & 26.9 & 24.5 & 23.1 & 30.3 & 52.5 & 42.5 & 47.4 \\
    FreeOcc-E (ours)                              & \textbf{43.6} & \textbf{36.3} & \textbf{32.2} & \textbf{28.9} & \textbf{26.5} & \textbf{24.9} & \textbf{32.1} & \textbf{54.6} & 45.5 & \textbf{49.6} \\
    \bottomrule
    \end{tabular}
    
    \vspace{3pt}
    \raggedright \scriptsize
    $^*$Indicates the models whose output point clouds we voxelize for evaluation.
    \vspace{-4mm}
\end{table}

\paragraph{Geometry forecasting (\cref{tab:geometry_forecasting}).}
On the point-based track, FreeOcc-E reaches the strongest NFCD at $\tau{=}6$ (0.47\,m vs.\ 0.58\,m for Copilot4D and 0.89\,m for UnO).
At $\tau{=}2$, Copilot4D remains best on NFCD (0.36 vs.\ 0.41), while FreeOcc-E is best on L1 depth error and AbsRel. The gap closes at longer horizons: Copilot4D rises from 0.36 to 0.58 NFCD, whereas FreeOcc-E rises from 0.41 to 0.47. FreeOcc-D achieves the highest recall (49.4\%) but struggles with false positives (41.3\% precision). FreeOcc-E uses uncertainty-aware blending to explicitly suppress noise, a visualization is provided in \Cref{fig:appx_ghost_1290}. This trade-off slightly lowers recall but boosts precision to 54.6\%, securing the best F1 score. FreeOcc-E also improves over FreeOcc-G, showing that uncertainty-aware blending is stronger than either branch alone.

\paragraph{Occupancy forecasting (\cref{tab:occupancy_forecasting}).}
On the voxel-based track, FreeOcc-E achieves the highest average mIoU at 32.1\%, improving over OccWorld by +3.2 points and over FreeOcc-G by +1.8 points. The ensemble is best at every horizon, reaching 24.9\% mIoU at $\tau{=}6$. FreeOcc-G consistently improves over FreeOcc-D, showing the benefit of continuous generation. The additional gain from FreeOcc-E indicates that uncertainty-aware blending balances the decoder's recall with the ODE's sharper long-horizon structure. Static Copy and 4D-Occ degrade much faster, and ViDAR remains limited by the point-to-voxel conversion used for dense occupancy evaluation.

\subsection{Efficiency Comparison}
\label{subsec:efficiency}

\Cref{tab:efficiency} shows that FreeOcc keeps a strong accuracy--speed trade-off. FreeOcc-D runs at 45.4 FPS, FreeOcc-E sustains 27.0 FPS while improving accuracy over all baselines, and FreeOcc-G remains compact but slower because of iterative ODE integration. Detailed latency analysis under adverse conditions is deferred to ~\cref{sec:appx_latency}.

\begin{table}[ht]
    \centering
    \caption{Efficiency comparison on nuScenes. We evaluate the parameter count and inference speed of our three variants: D (Decoder-only), G (Generation-only), and E (Ensemble) against representative baselines. All measurements are conducted on a single NVIDIA RTX 4090.}
    \label{tab:efficiency}
    \small
    \begin{tabular}{l c c c c}
    \toprule
    Method & \#Params & Latency (ms) & FPS \\
    \midrule
    4D-Occ~\cite{khurana2023point}    & 8.59M  & 73  & 13.7  \\
    OccWorld~\cite{zheng2025occworld} & 72.4M  & 56  & 17.8  \\
    ViDAR~\cite{yang2023vidar}          & 69.1M   & 189 & 5.3   \\
    \midrule
    FreeOcc-D (ours)                  & 20.7M  & 22  & 45.4  \\
    FreeOcc-G (ours)                  & 7.5M  & 168 & 6.0   \\
    FreeOcc-E (ours)                  & 28.3M  & 37  & 27.0  \\
    \bottomrule
    \end{tabular}
\end{table}
\subsection{Zero-Shot Cross-Domain Generalisation}
\label{subsec:zeroshot}

\Cref{tab:zeroshot} evaluates cross-domain generalisation by deploying nuScenes-trained models directly on SemanticKITTI with no fine-tuning.
Despite the substantial domain gap (different LiDAR beam count, sensor mounting height, and geographic driving context), FreeOcc-E remains clearly ahead of the baselines, outperforming OccWorld by +5.8 points at short term and +5.9 points at long term.
The gap is larger at long horizons, where FreeOcc-E reaches 18.2\% mIoU versus 12.3\% for OccWorld and 8.9\% for 4D-Occ. FreeOcc-E also improves over FreeOcc-G, again indicating that the ensemble is more robust than either branch alone under domain shift. Notably, while FreeOcc-D's aggressive generation yields higher recall at long horizons (33.2\% compared to 30.8\% for FreeOcc-E), it suffers from reduced precision. Additional SemanticKITTI visualizations are provided in ~\cref{sec:appx_kitti}.

\begin{table}[ht]
    \centering
    \caption{Zero-shot transfer from nuScenes to SemanticKITTI (no fine-tuning). We report metrics across short-term and long-term horizons to assess structural preservation under distribution shift.}
    \label{tab:zeroshot}
    \small
    \begin{tabular}{l | c c c | c c c}
    \toprule
    \multirow{2}{*}{Method} & \multicolumn{3}{c|}{Short-term ($\tau \leq 3$)} & \multicolumn{3}{c}{Long-term ($\tau > 3$)} \\
    \cmidrule{2-7}
     & mIoU & Precision & Recall & mIoU & Precision & Recall \\
    \midrule
    Static Copy                                & 14.5   & 30.2        & 22.1     & 7.5   & 22.5        & 13.0 \\
    4D-Occ~\cite{khurana2023point}             & 15.2   & 31.8        & 24.4     & 8.9   & 24.1        & 15.2 \\
    OccWorld~\cite{zheng2025occworld}          & 18.5   & 38.1        & 30.1     & 12.3   & 31.5        & 22.1 \\
    ViDAR~\cite{yang2023vidar}          & 6.5   & 12.4        & 8.3     & 2.3   & 1.5        & 2.2 \\
    FreeOcc-D (ours)                           & 22.0   & 42.5        & \textbf{39.8}     & 15.5   & 34.1        & \textbf{33.2} \\
    FreeOcc-G (ours)                           & 23.5   & 44.0        & 36.5     & 17.2   & 38.5        & 29.8 \\
    FreeOcc-E (ours)                           & \textbf{24.3} & \textbf{45.0} & 37.5 & \textbf{18.2} & \textbf{39.5} & 30.8 \\
    \bottomrule
    \end{tabular}
\end{table}

\subsection{Robustness and Ablation Summary}
\label{subsec:robustness_ablation}

Under timestamp jitter and 30\% history dropout, FreeOcc-E maintains a 29.8\% mIoU, compared with 19.0\% for OccWorld, 14.2\% for 4D-Occ, and 12.8\% for ViDAR. In ablations, the largest single drop comes from removing ego-motion warping (32.1$\to$25.8 mIoU), followed by removing flow matching (32.1\%$\to$28.4\% mIoU). FreeOcc-E also remains stronger than FreeOcc-G, confirming that uncertainty-aware blending adds measurable gains beyond the ODE alone. Full quantitative results are reported in ~\cref{sec:appx_additional_experiments}.

\subsection{Qualitative Results}
\label{subsec:qualitative}

\begin{figure*}[t]
    \centering
    \includegraphics[width=\linewidth,trim=0mm 4mm 0mm 70mm, clip]{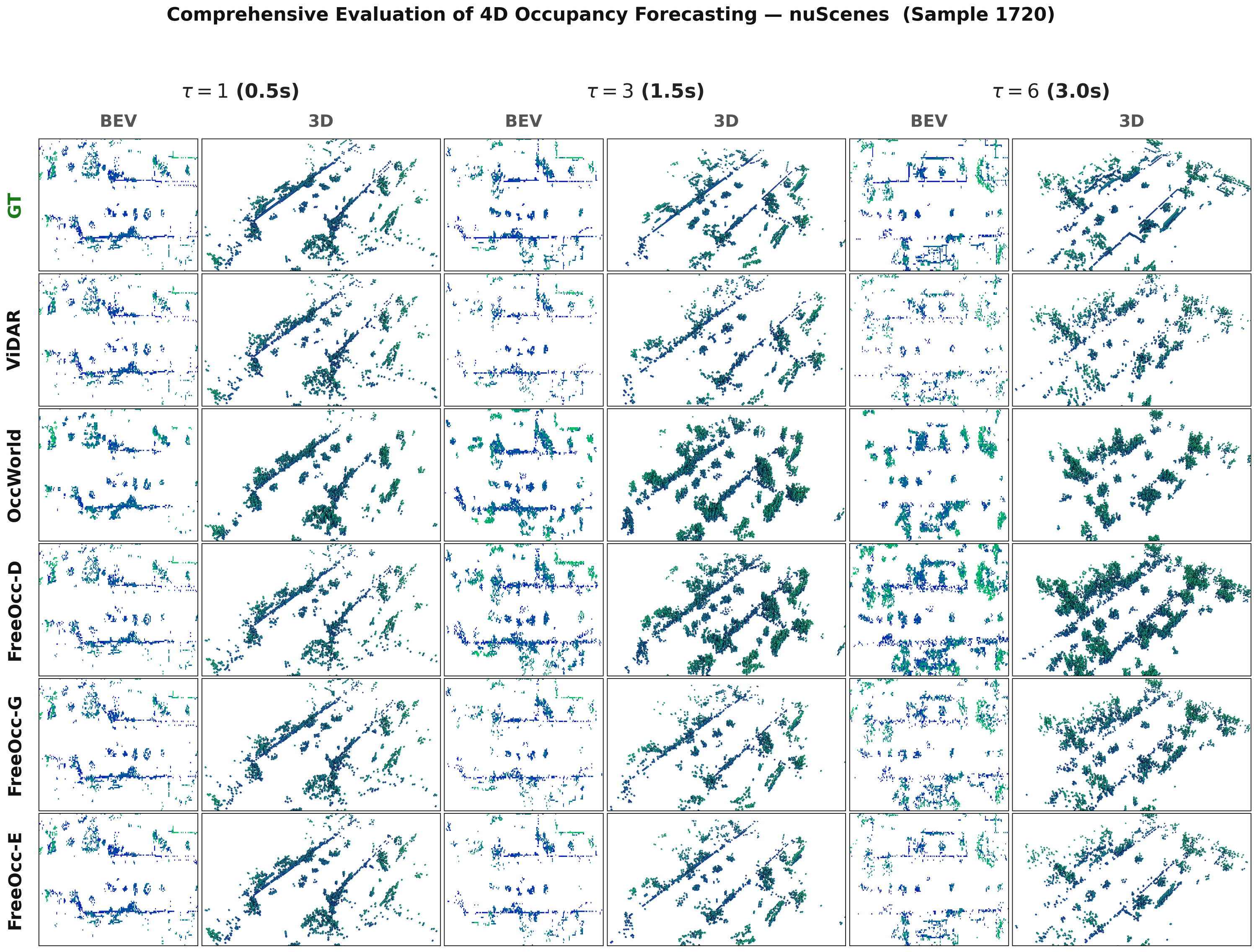}
    \caption{4D occupancy forecasting on nuScenes(sample 1720) in BEV and 3D views across six horizons. FreeOcc-E is compared against OccWorld and ViDAR; more qualitative examples are given in ~\cref{sec:appx_qualitative}.}
    \label{fig:qualitative_main}
    \vspace{-4mm}
\end{figure*}

\Cref{fig:qualitative_main} shows the qualitative trend behind the tables. FreeOcc-E preserves cleaner occupancy boundaries and more stable long-horizon structure than OccWorld and ViDAR, especially in dynamic and far-field regions. Additional visualizations, uncertainty maps, and robustness examples can be found in ~\cref{sec:appx_qualitative}--\cref{sec:appx_robustness}.

\section{Conclusion}

We introduce FreeOcc, a continuous-time world model for 4D occupancy forecasting. By explicitly decoupling ego-motion via SE(2) warping, we isolate residual scene dynamics. This formulation enables attention-free temporal fusion and physics-constrained continuous flow matching. To reduce inference latency, an uncertainty-aware routing mechanism processes well-observed geometry with a single-pass decoder, reserving ODE integration strictly for ambiguous regions. Operating with roughly 40\% of the parameters of comparable models, FreeOcc achieves 32.1\% mIoU on nuScenes at 27 FPS. Furthermore, the model demonstrates zero-shot transfer to SemanticKITTI and retains 92.8\% of its mIoU under simulated sensor degradation. These results indicate that explicit geometric decoupling and physical priors provide an effective, robust foundation for autonomous driving perception.

Limitations and future directions are discussed in ~\cref{sec:appx_limitations}.

\section*{Acknowledgements}
This work was partially funded by the Natural Sciences and Engineering Research Council (NSERC) of Canada.
\copyright~2026 His Majesty the King in Right of Canada, as represented by the National Research Council of Canada.


\medskip

\bibliographystyle{plainnat}
\bibliography{reference}


\appendix

\section{Additional Method Details}
\label{sec:appx_method_details}



\subsection{Temporal augmentation and decoder details}

\paragraph{Range encoding.}
\label{sec:appx_range_encoding}
For BEV cell $(h,w)$ with radial distance $r_{hw}=\sqrt{x_h^2+y_w^2}$, we use
\begin{equation}
    \label{eq:range_enc}
    \rho_{hw} = W_2\;\mathrm{ReLU}\!\bigl(W_1\,[\sin(2^0 r),\cos(2^0 r),\dots,\sin(2^{B_r-1} r),\cos(2^{B_r-1} r)]\bigr),
\end{equation}
and add it to the pillar pseudo-image as $\bar{F}_i\leftarrow\bar{F}_i+\rho$.

\paragraph{Timestamp jitter and sensor dropout.}
\label{sec:appx_augmentation}
To train the temporal aggregator on irregular input streams, each history sweep $P_i$ ($i<t$) is assigned a perturbed offset
\begin{equation}
    \label{eq:jitter}
    \Delta t_i' = \Delta t_i + \eta_i, \qquad \eta_i \sim \mathcal{U}(-\delta_t,\delta_t),
\end{equation}
with $\delta_t=0.1$\,s. The current frame is never perturbed. In parallel, every non-current sweep is dropped independently with probability $p_\mathrm{drop}=0.3$, while forcing at least one earlier sweep to remain. Jitter trains continuous interpolation over residual timing noise, whereas dropout trains recovery from incomplete histories.

The timestamp embedding itself uses exponentially spaced sinusoidal frequencies:
\begin{equation}
    \label{eq:time_enc}
    \phi(\Delta t_i) = \bigl[\sin(\omega_1\Delta t_i),\;\cos(\omega_1\Delta t_i),\;\dots,\;\sin(\omega_{C/2}\Delta t_i),\;\cos(\omega_{C/2}\Delta t_i)\bigr],
\end{equation}
where $\omega_j=\exp(-j\ln\Lambda/(C/2))$.

\paragraph{Decoder architecture.}
\label{sec:appx_decoder}
Each BEV prior $C_\tau$ is refined by two $3\!\times\!3$ BEV convolutions, lifted to $C\!\cdot\!Z$ channels by a $1\!\times\!1$ convolution, reshaped to $\mathbb{R}^{C\times H\times W\times Z}$, augmented with a height positional encoding $z_\mathrm{pos}$, and processed by two $3\!\times\!3\!\times\!3$ volume convolutions. A future-step embedding $e_\tau$ is added before lifting. The occupancy head maps $C\!\to\!C/2\!\to\!C/4\!\to\!1$, while the uncertainty head predicts $\hat{s}_\tau$ with a two-layer MLP whose bias is initialised to $-0.5$.

\paragraph{Velocity network details.}
\label{sec:appx_flow_network}
The velocity network $v_\theta$ is a 3D U-Net with channel widths $\{c,2c,4c\}$, a bottleneck, and symmetric skip connections. Flow time $s$ is injected with a sinusoidal embedding, and each future step $\tau$ has a learned embedding. For diversity, a fraction $\alpha=0.2$ of initial Gaussian samples is replaced by slightly noised decoder outputs.

\paragraph{Uncertainty normalisation for cooperative flow matching.}
\label{sec:appx_uncertainty_norm}
The normalisation bounds $\mu_{\mathrm{low}}$ and $\mu_{\mathrm{high}}$ in \cref{eq:cfm_loss} are maintained as exponential moving averages of the 5th and 95th percentiles of detached $\hat{s}_v$ computed over training batches. This ensures that the uncertainty weight $w$ remains stable while effectively spending generative capacity strictly in regions where the discriminative path is least reliable.

\subsection{Full loss definitions}
\label{sec:appx_loss_defs}

\paragraph{Focal BCE $\mathcal{L}_\mathrm{occ}$~\cite{lin2017focal}.}
We use focal BCE with $\alpha=0.75$ to upweight the minority occupied class and suppress easy free-space gradients.

\paragraph{Lov\'asz hinge $\mathcal{L}_\mathrm{Lov}$~\cite{berman2018lovasz}.}
This term is a differentiable surrogate for IoU. Its weight is warmed up as
\begin{equation}
    \lambda_\mathrm{lov}(n) = \bar{\lambda}_\mathrm{lov}\cdot\min\!\left(1, \max\!\left(0, \frac{n-n_\mathrm{start}}{n_\mathrm{ramp}}\right)\right).
\end{equation}

\paragraph{Dice $\mathcal{L}_\mathrm{Dice}$~\cite{milletari2016vnet}.}
Dice complements BCE and Lov\'asz with a smooth overlap objective.

\paragraph{Sparsity $\mathcal{L}_\mathrm{sp}$.}
We penalise deviation from a target occupied ratio $\bar{r}=0.05$ using $|\bar{\sigma}(\hat{y})-\bar{r}|$, with a weight that grows linearly with horizon as $(1+0.5\tau)$.

\paragraph{Temporal consistency $\mathcal{L}_\mathrm{tc}$.}
We match predicted and ground-truth temporal derivatives:
\begin{equation}
    \mathcal{L}_\mathrm{tc}=\frac{1}{N-1}\sum_\tau\left\|\bigl(\sigma(\hat{y}_{\tau+1})-\sigma(\hat{y}_\tau)\bigr)-\bigl(y_{\tau+1}-y_\tau\bigr)\right\|_2^2.
\end{equation}

\paragraph{Ray-traced free-space $\mathcal{L}_\mathrm{ray}$.}
LiDAR rays define observed free space before the first hit. With traversed voxel set $\mathcal{F}$, we use
\begin{equation}
    \mathcal{L}_\mathrm{ray} = \frac{1}{|\mathcal{F}|}\sum_{(i,j,k)\in\mathcal{F}} \sigma(\hat{y}_{ijk}).
\end{equation}
Voxels behind the first surface hit remain unconstrained.

\paragraph{Conservation $\mathcal{L}_\mathrm{cons}$.}
Under a closed-world approximation, total occupied mass should vary smoothly:
\begin{equation}
    \mathcal{L}_\mathrm{cons} = \frac{1}{N-1}\sum_\tau\bigl(\Delta\hat{M}_\tau-\Delta M_\tau^{\mathrm{gt}}\bigr)^2,
\end{equation}
where $\Delta\hat{M}_\tau$ is measured over the interior region $\Omega$.

\paragraph{Range-weighted loss $\mathcal{L}_\mathrm{rw}$.}
We modulate supervision by sensor distance using
\begin{equation}
    w(r)=w_\mathrm{near}(1-r/r_\mathrm{max})+w_\mathrm{far}\,r/r_\mathrm{max},
\end{equation}
with $w_\mathrm{near}=1.5$ and $w_\mathrm{far}=0.5$.

\paragraph{Velocity smoothness $\mathcal{L}_\mathrm{vel}$.}
~\cref{eq:vel_smooth} imposes piecewise-smooth motion on the ego-decoupled flow field.

\paragraph{Uncertainty calibration $\mathcal{L}_\mathrm{unc}$.}
Following the heteroscedastic classification approach, we supervise the uncertainty head using:
\begin{equation}
    \mathcal{L}_\mathrm{unc} = \frac{1}{|\Omega|}\sum_{ijk}\left[\exp(-\hat{s}_{ijk})\,\ell_{\text{BCE}}(\hat{y}_{ijk}, y_{ijk}) + \beta\hat{s}_{ijk}\right],
\end{equation}
where $\ell_{\text{BCE}}$ denotes the binary cross-entropy loss applied to the predicted occupancy logits $\hat{y}_{ijk}$. We fix $\beta=0.5$ as a penalty to prevent the model from trivializing the objective by predicting infinite variance.

\paragraph{Hyperparameter configurations.} 
For the composite objective defined in Eq.~\ref{eq:total_loss}, the base scalar weights are empirically set as follows: $\lambda_\mathrm{occ}=2.0$, $\lambda_\mathrm{lov}=1.0$, $\lambda_\mathrm{dice}=1.0$, $\lambda_\mathrm{sp}=0.5$, $\lambda_\mathrm{tc}=1.0$, $\lambda_\mathrm{ray}=1.0$, $\lambda_\mathrm{cons}=0.1$, $\lambda_\mathrm{vel}=0.5$, $\lambda_\mathrm{rw}=1.0$, and $\lambda_\mathrm{unc}=1.0$. For the range encoding (Eq.~\ref{eq:range_enc}), the frequency band limit is $B_r = 6$, and for the timestamp embedding (Eq.~\ref{eq:time_enc}), the base period $\Lambda = 10000$. The scheduling parameters for the Lov\'asz hinge warmup are $n_\mathrm{start}=1000$ and $n_\mathrm{ramp}=2000$ iterations.

\section{Limitations and Future Directions}
\label{sec:appx_limitations}

FreeOcc does not model semantic classes; extending the framework to class-conditional flow matching remains future work.
The ODE branch still adds non-trivial inference cost, which motivates future work on adaptive solvers or distillation.

Furthermore, the explicit geometric decoupling relies on a planar SE(2) ego-motion warp. This assumes flat-world dynamics, which may introduce minor spatial misalignments under severe pitch, roll, or steep inclines. However, enforcing SE(2) decoupling rather than a full 3D volumetric SE(3) warp drastically reduces the computational overhead of grid sampling. In practice, our evaluations demonstrate that the residual ConvGRU effectively absorbs minor Z-axis translations and pitch deviations, confirming that SE(2) alignment strikes the optimal balance between geometric accuracy and real-time inference latency for urban driving scenarios. Furthermore, the model treats the motion planner's trajectory as a known input; however, experiments show that a self-predicted trajectory extrapolated from the vehicle's recent speed and yaw rate, using no future information, still achieves 31.2\% mIoU (vs.\ 32.1\% with the recorded route), indicating that the method does not critically depend on an accurate external plan. Sensitivity to trajectory noise is analysed in \cref{sec:appx_traj_robustness}. Mitigating vulnerability to local odometry or planning drift remains an area for future refinement.

Finally, the point-based track retains a quantisation floor induced by the voxel resolution ($\delta\!=\!0.4$\,m). Since $\delta$ is a preprocessing choice rather than an architectural constraint, finer grids or a learned sub-voxel refinement head could reduce this error floor at additional compute cost.

\section{Experimental Details}
\label{sec:appx_metrics}

\subsection{Dataset and transfer details}

We train on the standard nuScenes split (700/150/150 train/val/test) with 32-beam LiDAR, 20\,Hz sweeps, and 2\,Hz keyframes. For zero-shot transfer, we evaluate on SemanticKITTI. Before evaluation, SemanticKITTI point clouds are range-clipped to $[-40,40]$\,m in the horizontal plane, shifted by $0.11$\,m along $Z$, and beam-downsampled to reduce the 64-to-32 line density mismatch.

\subsection{Metric Alignment with Baselines}
\label{sec:metric_alignment}
In the point-based evaluation track (\cref{tab:geometry_forecasting}), we report Near Field Chamfer Distance (NFCD) to align with UnO~\cite{agro2024uno}.

Copilot4D~\cite{zhang2023learning} does not use the term ``NFCD'', but it reports Chamfer Distance within a fixed Region of Interest (ROI). This evaluation protocol is equivalent to the NFCD computation used by UnO. We therefore report Copilot4D's ROI-restricted Chamfer Distance in the NFCD column.

\subsection{Discussion: Sensor-Grounded vs.\ Completion-Based Evaluation}
\label{sec:appx_occ3d}

We do not evaluate on Occ3D~\cite{tian2024occ3d} because it measures a different task. Occ3D targets \emph{occupancy completion}: reconstructing the full 3D scene, including regions never observed by LiDAR, such as building interiors and heavily occluded surfaces. FreeOcc instead addresses \emph{sensor-grounded occupancy forecasting}: predicting the 3D structure that the sensor will observe at future timesteps.

This distinction matters for planning. The relevant target is future sensor-observable geometry, not the interior of occluded structures. Accordingly, our supervision is derived from future LiDAR returns rather than dense completion labels. Training against completion labels would force the model to infer geometry behind walls and inside objects, which is underdetermined and unnecessary for the forecasting setting considered here.

The evaluation objectives also differ. Occ3D rewards filling unobserved regions, whereas our protocol measures binary occupied/free IoU against future scans and therefore rewards accurate prediction of observable structure. These objectives need not align: a model tuned for completion may hallucinate geometry that improves completion scores but is undesirable for downstream planning.

\subsection{Additional experimental results}
\label{sec:appx_additional_experiments}

\subsubsection{Robustness to sensor temporal imperfections}
\label{sec:appx_additional_experiments_Robustness}
\begin{table}[!htbp] 
\centering
\caption{Robustness to sensor temporal imperfections on nuScenes. The imperfect setting adds severe timestamp jitter and 30\% history-frame dropout. All metrics are voxel-based.}
\label{tab:app_sensor_robustness}
\small
\setlength{\tabcolsep}{4pt}
\begin{tabular}{l l c c c c}
\toprule
Method & Condition & mIoU $\uparrow$ & Prec. $\uparrow$ & Rec. $\uparrow$ & F1 $\uparrow$ \\
\midrule
\multirow{2}{*}{4D-Occ~\cite{khurana2023point}}
& Ideal     & 22.6 & 39.1 & 31.8 & 35.1 \\
& Imperfect & 14.2 & 30.5 & 21.0 & 24.9 \\
\midrule
\multirow{2}{*}{OccWorld~\cite{zheng2025occworld}}
& Ideal     & 28.9 & 50.5 & 40.2 & 44.8 \\
& Imperfect & 19.0 & 38.5 & 28.5 & 32.8 \\
\midrule
\multirow{2}{*}{ViDAR~\cite{yang2023vidar}}
& Ideal     & 20.5 & 39.5 & 29.8 & 34.1 \\
& Imperfect & 12.8 & 28.2 & 19.5 & 23.1 \\
\midrule
\multirow{2}{*}{FreeOcc-D (ours)}
& Ideal     & 28.4 & 41.3 & \textbf{49.4} & 44.9 \\
& Imperfect & 26.0 & 37.5 & \textbf{46.5} & 41.3 \\
\midrule
\multirow{2}{*}{FreeOcc-G (ours)}
& Ideal     & 30.3 & 52.5 & 42.5 & 47.4 \\
& Imperfect & 28.1 & 49.8 & 39.0 & 43.7 \\
\midrule
\multirow{2}{*}{FreeOcc-E (ours)}
& Ideal     & \textbf{32.1} & \textbf{54.6} & 45.5 & \textbf{49.6} \\
& Imperfect & \textbf{29.8} & \textbf{51.5} & 42.0 & \textbf{46.3} \\
\bottomrule
\end{tabular}
\end{table}

FreeOcc retains the strongest performance under temporal noise. FreeOcc-E keeps 92.8\% of its ideal mIoU (29.8/32.1), while OccWorld retains 65.7\%, 4D-Occ 62.8\%, and ViDAR 62.4\%. This is consistent with the timestamp-jitter and sensor-dropout training described in \cref{subsec:aggregator}. Detailed results are shown in \cref{tab:app_ablation}.

\subsubsection{Ablation studies}

\begin{table}[!htbp]
    \centering
    \caption{Ablation study on nuScenes validation. Metrics are voxel-based. The results demonstrate that explicit geometric decoupling and physics-aware regularization are the primary drivers of forecasting quality.}
    \label{tab:app_ablation}
    \footnotesize
    \setlength{\tabcolsep}{4pt}
    \begin{tabular}{p{0.56\columnwidth} c c c}
    \toprule
    Variant & mIoU $\uparrow$ & Precision $\uparrow$ & F1 $\uparrow$ \\
    \midrule
    FreeOcc-E (full) & 32.1 & 54.6 & 49.6 \\
    \midrule
    \multicolumn{4}{l}{\textit{Generation Path Ablations}} \\
    \quad w/o Flow Matching (FreeOcc-D) & 28.4 & 41.3 & 44.9 \\
    \quad w/o Decoder (FreeOcc-G) & 30.3 & 52.5 & 47.4 \\
    \quad Max blending & 31.2 & 53.4 & 48.4 \\
    \quad Mean blending & 30.5 & 52.6 & 47.5 \\
    \midrule
    \multicolumn{4}{l}{\textit{Architecture Ablations}} \\
    \quad w/o Ego-Motion Warping & 25.8 & 46.2 & 40.5 \\
    \quad w/o Residual Connection & 30.2 & 52.1 & 47.2 \\
    \quad w/o Delay-Aware Confidence Gate & 31.1 & 53.3 & 48.3 \\
    \quad w/o Sensor Dropout & 31.4 & 53.8 & 48.8 \\
    \quad w/o Timestamp Jitter ($\delta_t=0$) & 31.6 & 54.0 & 49.0 \\
    \quad Fixed-stride temporal encoding & 30.8 & 52.9 & 48.0 \\
    \midrule
    \multicolumn{4}{l}{\textit{Physics-Aware Regularization Ablations}} \\
    \quad w/o \textbf{Structural Fidelity} (Group) & 28.1 & 49.5 & 44.1 \\
    \quad w/o \textbf{Physical Priors} (Group) & 28.7 & 50.8 & 44.8 \\
    \quad w/o $\mathcal{L}_\mathrm{vel}$ (Smoothness) & 30.4 & 52.3 & 47.4 \\
    \quad w/o $\mathcal{L}_\mathrm{ray}$ (Free-space) & 31.1 & 53.2 & 48.3 \\
    \quad w/o $\mathcal{L}_\mathrm{cons}$ (Mass conservation) & 31.4 & 53.8 & 48.8 \\
    \quad w/o $\mathcal{L}_\mathrm{unc}$ (Calibration) & 31.5 & 53.9 & 48.9 \\
    \quad w/o $\mathcal{L}_\mathrm{rw}$ (Range weighting) & 31.6 & 54.1 & 49.0 \\
    \midrule
    \multicolumn{4}{l}{\textit{Uncertainty Ablations}} \\
    \quad w/o cooperative unc.-guided FM & 30.6 & 52.7 & 47.7 \\
    \quad w/o uncertainty-aware blending & 31.2 & 53.4 & 48.4 \\
    \bottomrule
    \end{tabular}
\end{table}

The ablations in \cref{tab:app_ablation} support our core design philosophy. Removing Ego-Motion Warping causes the single largest drop (32.1 $\to$ 25.8 mIoU), confirming that explicit geometric decoupling is fundamental to forecasting accuracy. Within our loss system, the Structural Fidelity group is essential for grounding the discriminative path, while the Physical Priors (specifically velocity smoothness $\mathcal{L}_\mathrm{vel}$) ensure that the generative ODE produces kinematically plausible trajectories. Notably, removing either entire regularization group leads to a performance drop significantly larger than removing any individual term, proving that our physics-aware objectives act as a cohesive system rather than a collection of independent penalties.

\newpage
\section{Additional Qualitative Results}
\label{sec:appx_qualitative}

This appendix provides additional visualizations supporting the qualitative analysis in \cref{subsec:qualitative}.

\begin{figure*}[!htbp]
    \centering
    \includegraphics[width=\linewidth,trim=0mm 0mm 0mm 60mm, clip]{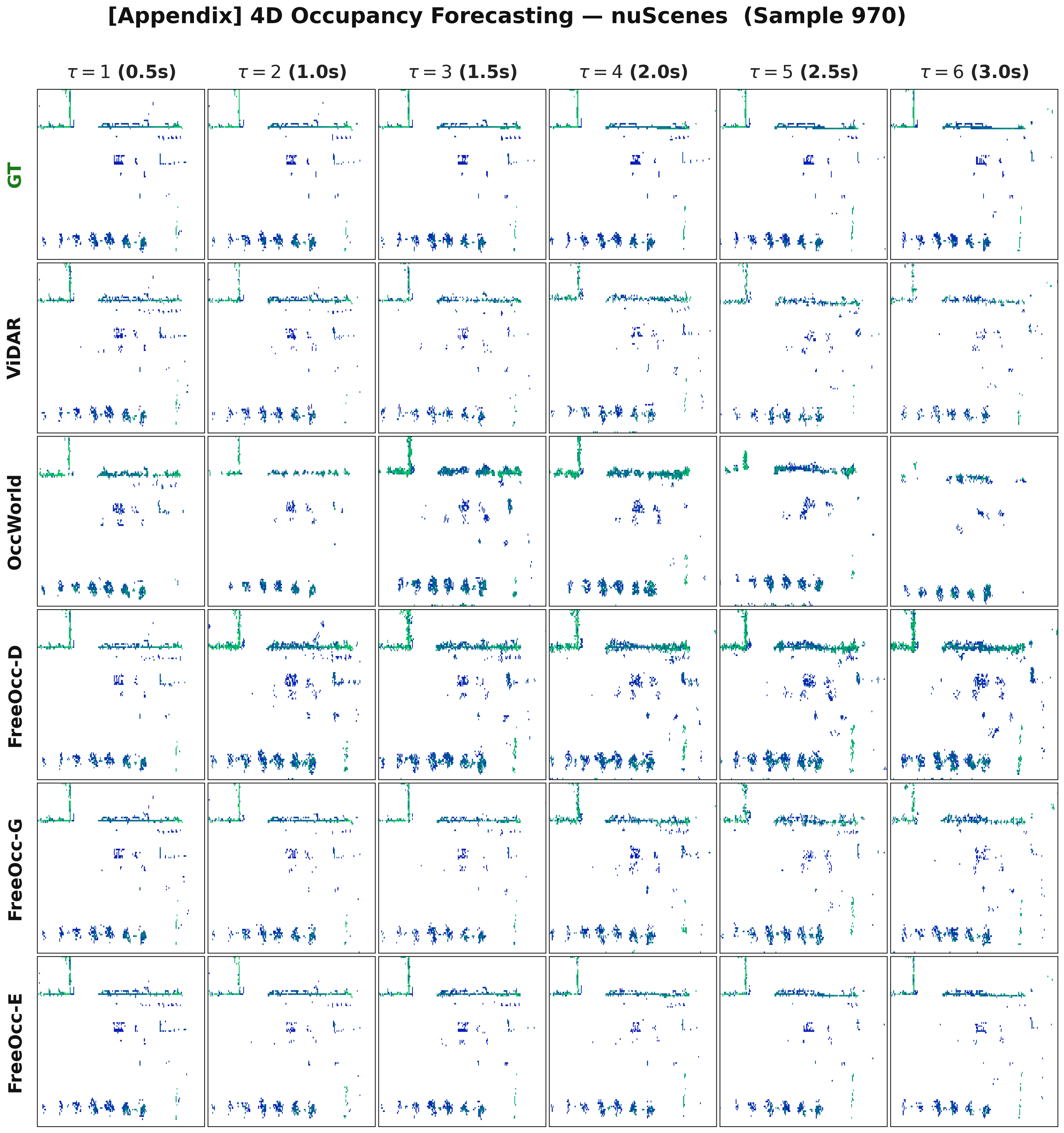}
    \caption{\textbf{BEV visualization of 4D occupancy forecasting (additional scene, sample 0970).}
    Columns span $\tau{=}1$ to $\tau{=}6$, comparing FreeOcc against OccWorld and ViDAR on a dense urban scene with multiple interacting vehicles.
    FreeOcc-E preserves scene layout and object extent more faithfully over the full 3.0\,s horizon, whereas OccWorld and ViDAR drift more noticeably after $\tau{=}3$.}
    \label{fig:appx_comparison_0970}
\end{figure*}

\begin{figure*}[!htbp]
    \centering
    \includegraphics[width=\linewidth,trim=0mm 0mm 0mm 25mm, clip]{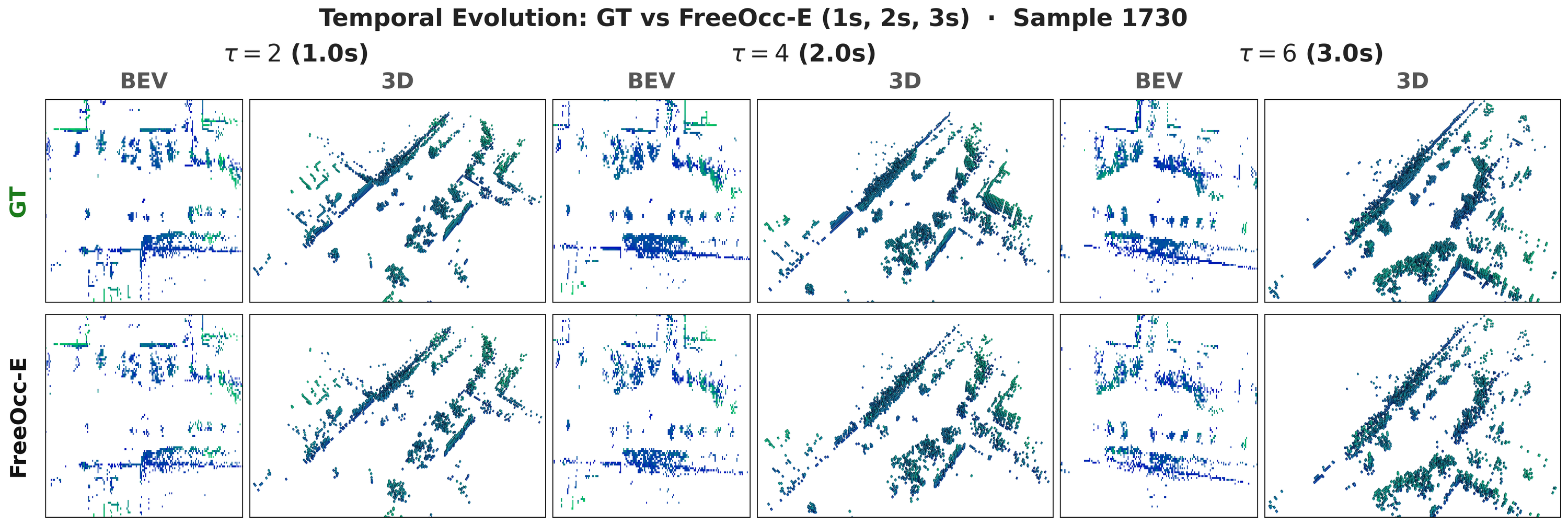}
    \caption{\textbf{Temporal evolution of GT vs.\ FreeOcc-E (nuScenes sample 1730).}
    Top row: ground-truth occupancy at $\tau{=}2$ (1.0\,s), $\tau{=}4$ (2.0\,s), and $\tau{=}6$ (3.0\,s), each shown as BEV (left) and 3D voxel rendering (right).
    Bottom row: corresponding FreeOcc-E predictions.
    Agreement is strongest at short horizons.
    By $\tau{=}6$, the main discrepancies appear in the far field and around dynamic actors, while large static structures remain well preserved.}
    \label{fig:appx_temporal_1730}
\end{figure*}

\begin{figure*}[!htbp]
    \centering
    \includegraphics[width=\linewidth,trim=0mm 0mm 0mm 25mm, clip]{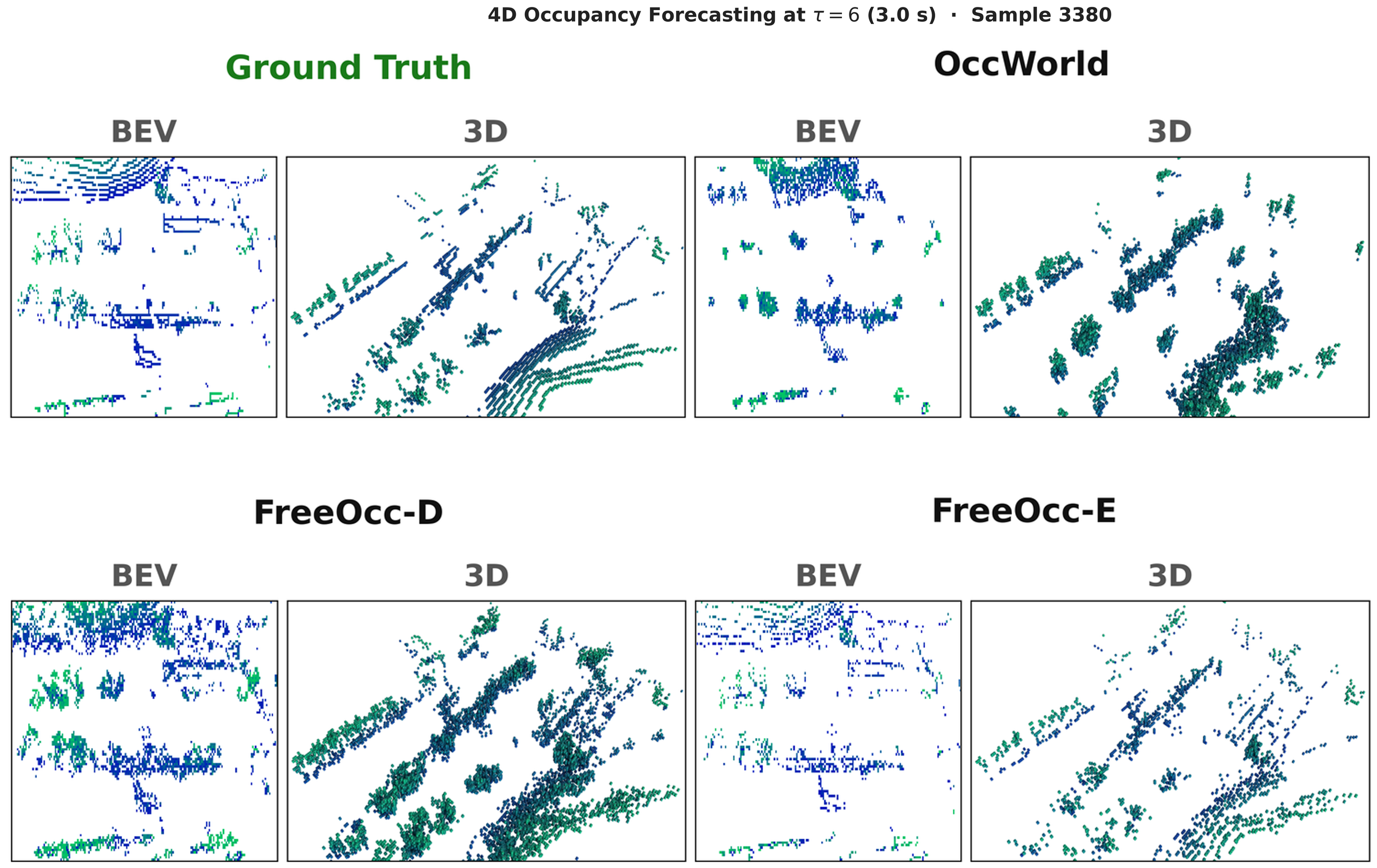}
    \caption{\textbf{Long-horizon comparison at $\tau{=}6$ (3.0\,s, nuScenes sample 3380).}
    Ground Truth, OccWorld, FreeOcc-D, and FreeOcc-E are shown in paired BEV and 3D views.
    At 3.0\,s, OccWorld exhibits boundary blur and noise around moving objects.
    FreeOcc-D slightly over-predicts occupancy, whereas FreeOcc-E preserves sharper boundaries and cleaner free space.}
    \label{fig:appx_longhorizon_3380}
\end{figure*}

\begin{figure*}[!htbp]
    \centering
    \begin{tabular}{@{}c@{\hspace{2pt}}c@{\hspace{2pt}}c@{}}
        \includegraphics[width=0.33\linewidth]{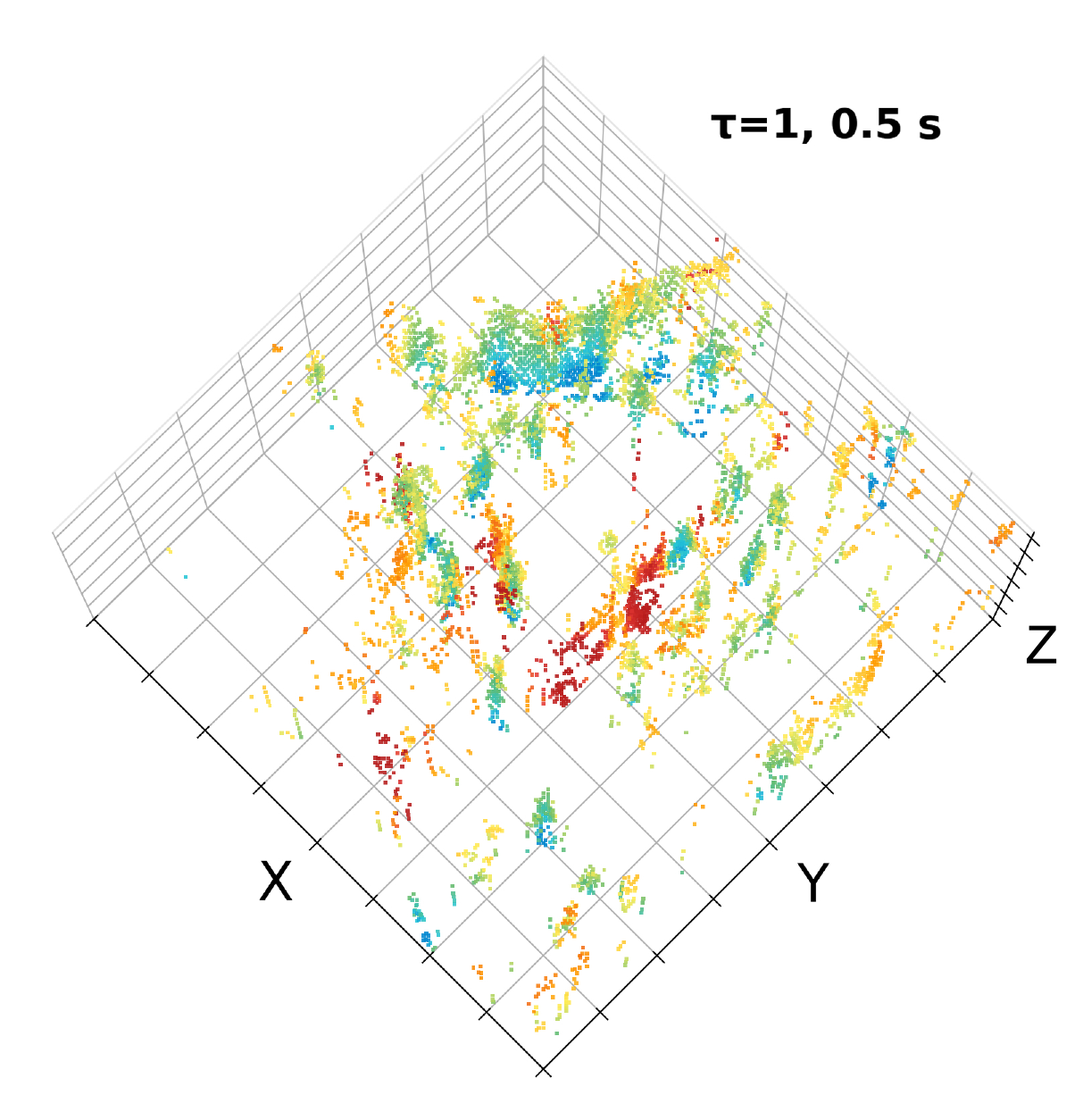} &
        \includegraphics[width=0.33\linewidth]{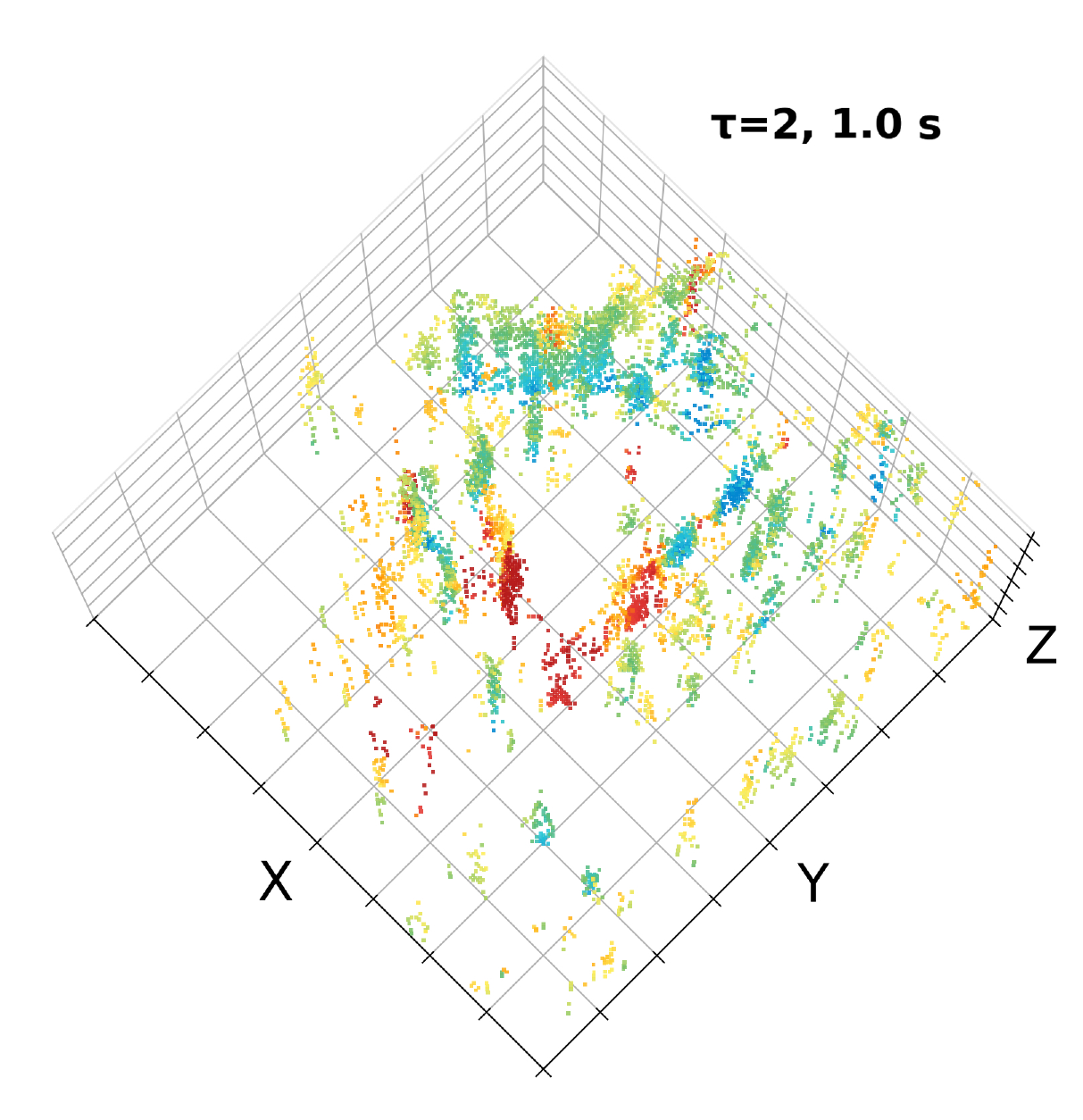} &
        \includegraphics[width=0.33\linewidth]{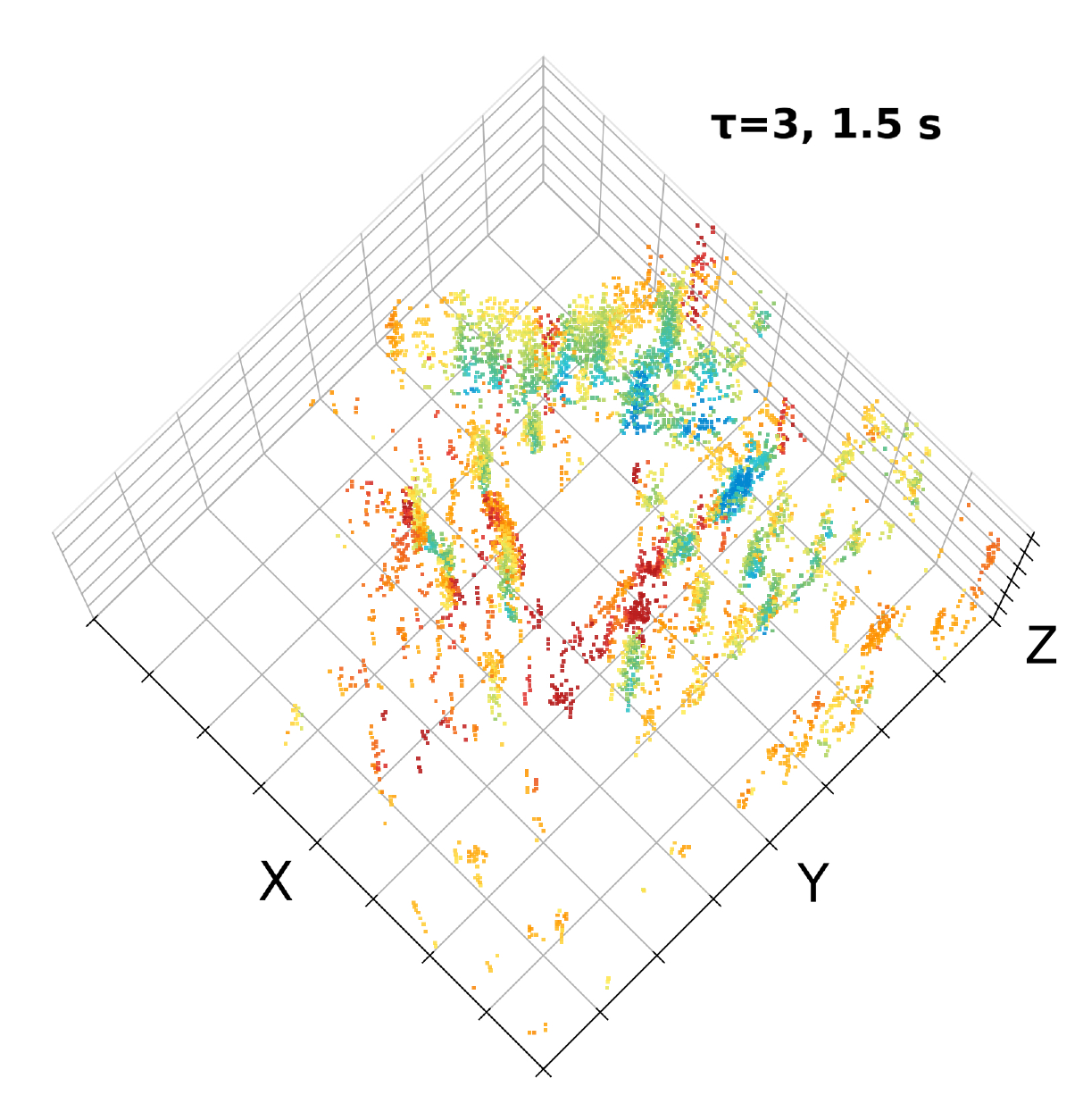} \\
        \includegraphics[width=0.33\linewidth]{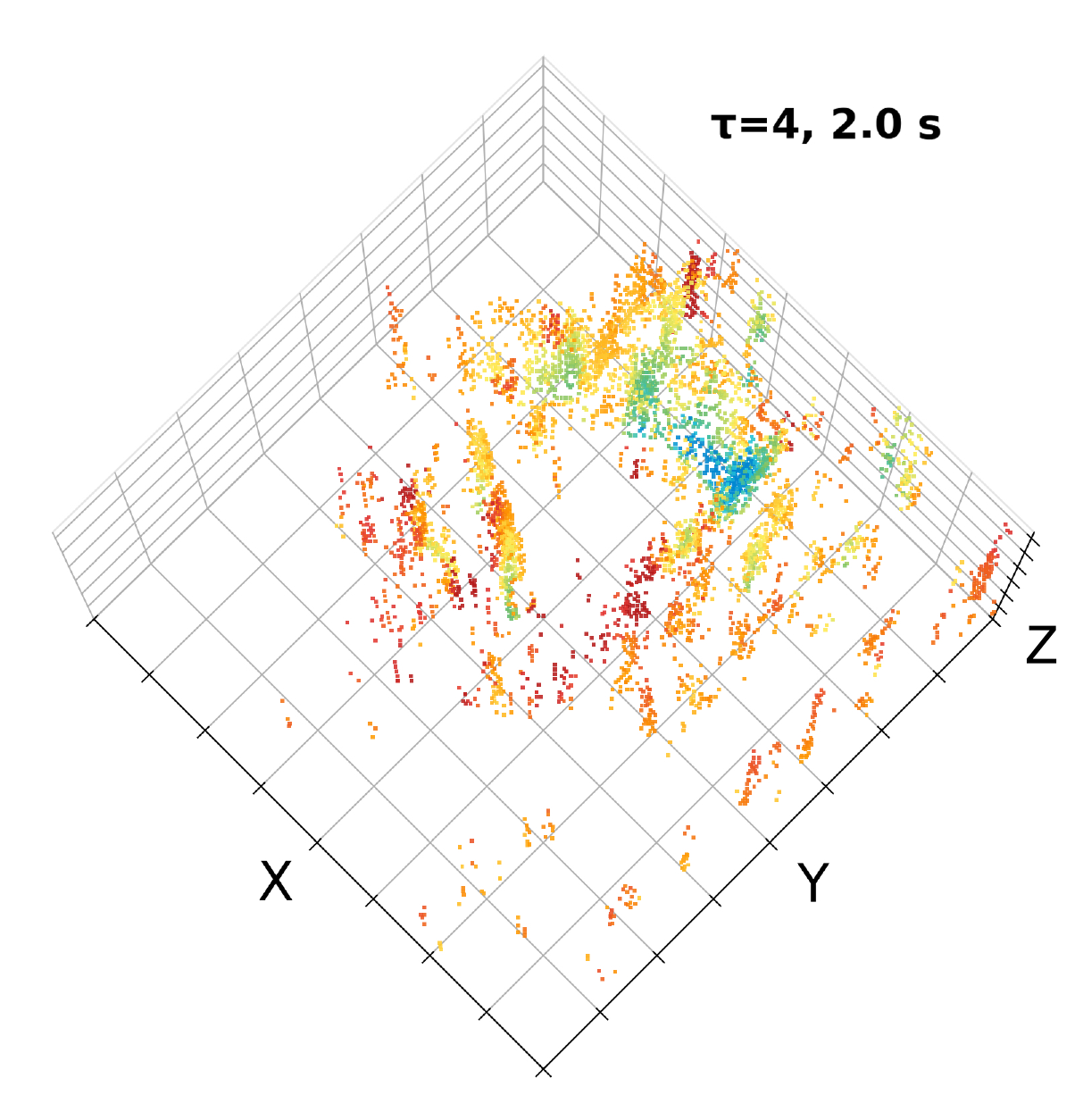} &
        \includegraphics[width=0.33\linewidth]{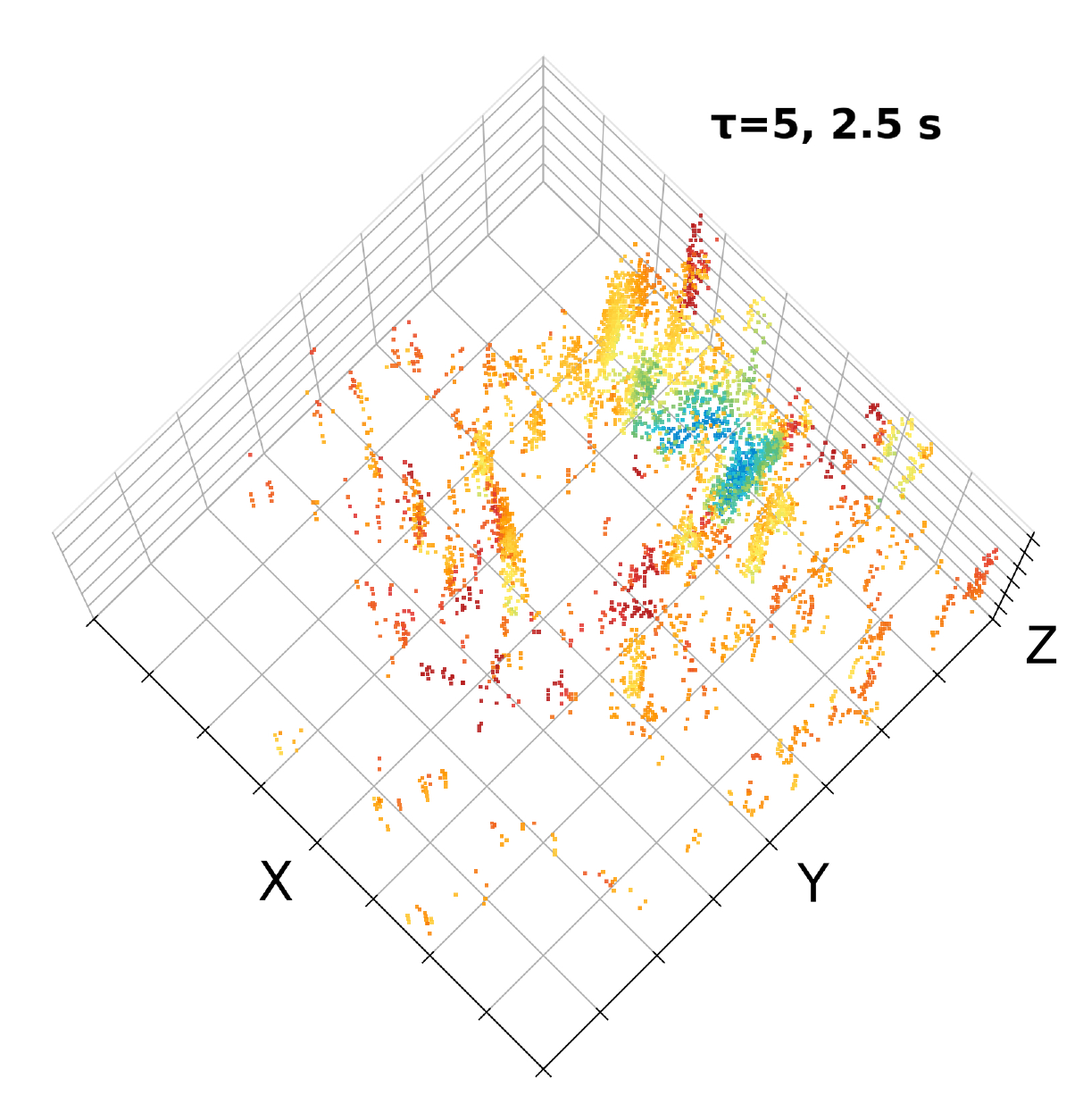} &
        \includegraphics[width=0.33\linewidth]{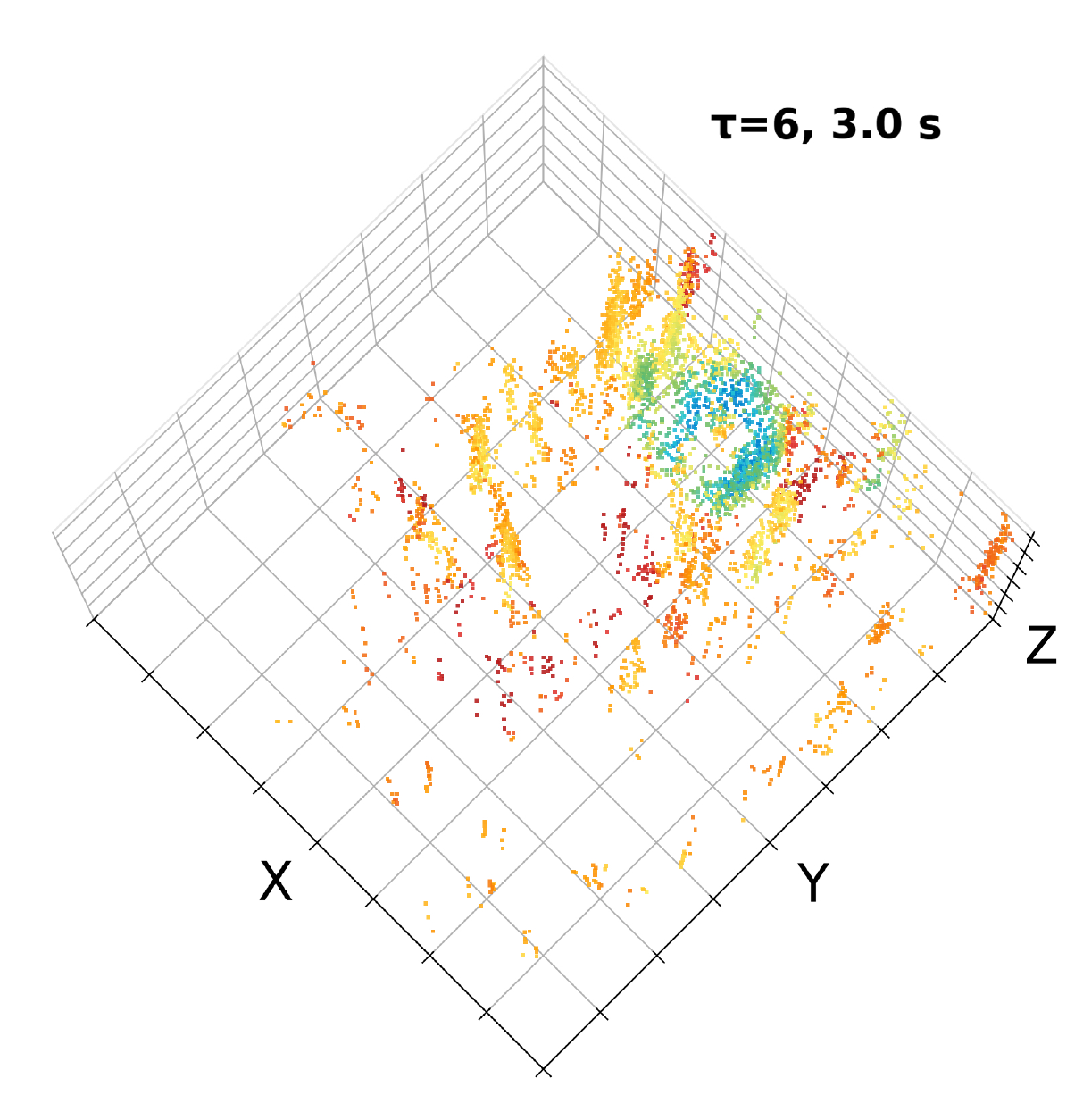} \\
    \end{tabular}
    \caption{\textbf{3D voxel-level uncertainty across prediction horizons (scene-0274, $\tau{=}1$ to $\tau{=}6$).}
    Each panel renders occupied voxels coloured by the decoder's predicted aleatoric uncertainty~$\hat{s}_v$; cool tones indicate low uncertainty and warm tones indicate high.
    At short horizons, uncertainty is concentrated near dynamic actors and occupancy boundaries.
    As the horizon increases, uncertainty expands into the far field and intensifies around moving objects.
    By $\tau{=}6$ (3.0\,s), most remaining structures are assigned substantially higher uncertainty.
    This behaviour is consistent with the ensemble design in \cref{subsec:blending}, which assigns more weight to the ODE path where decoder confidence weakens.}
    \label{fig:appx_scatter_unc}
\end{figure*}

\begin{figure*}[!htbp]
    \centering
    \includegraphics[width=0.7\linewidth,trim=0mm 0mm 30mm 0mm, clip]{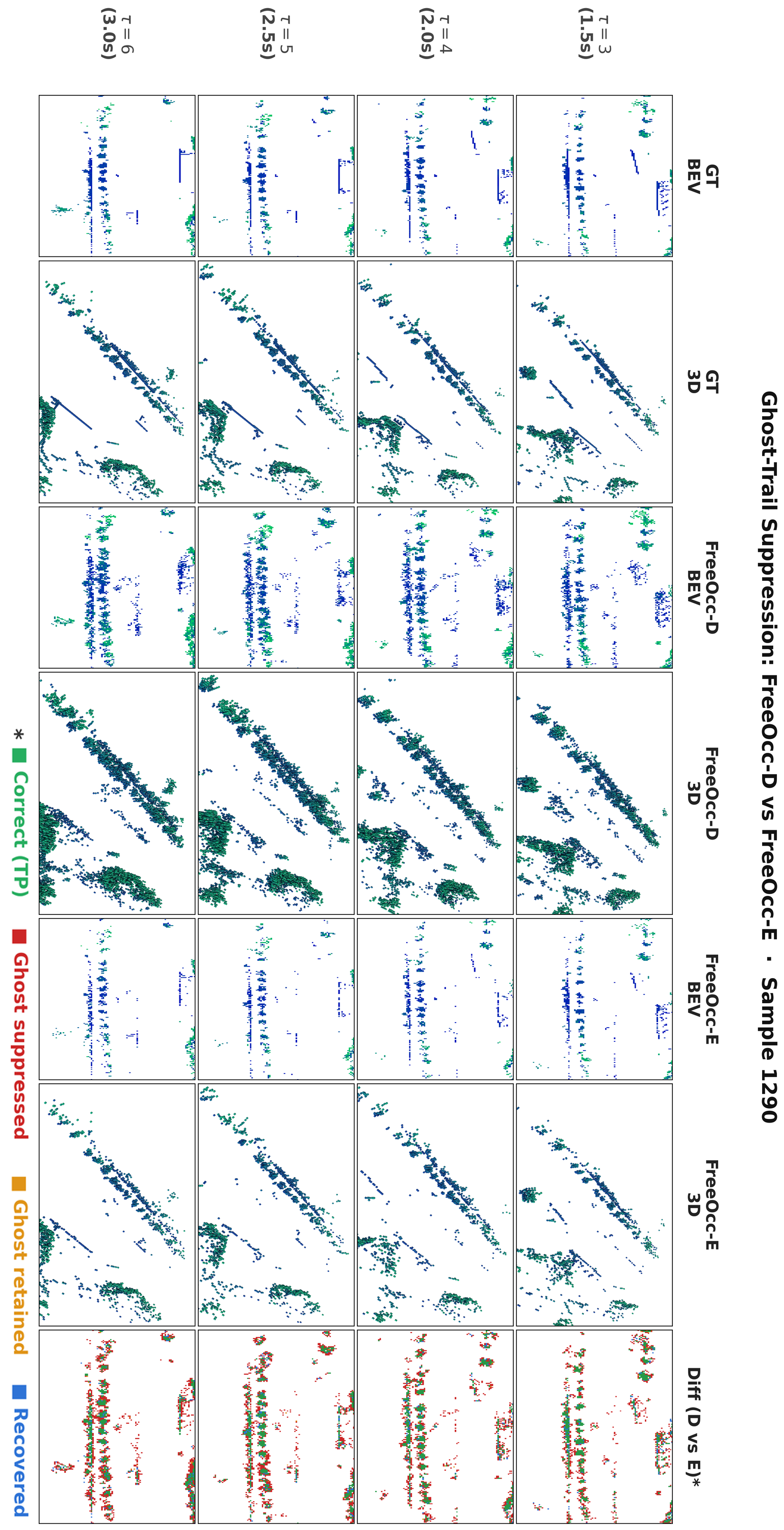}
    \caption{\textbf{Noise suppression analysis (nuScenes sample 1290).}
    Paired BEV and 3D views highlighting hallucinated occupancy at long prediction horizons between FreeOcc-D and FreeOcc-E.
    FreeOcc-E clears vacated space more effectively while preserving plausible future object positions.}
    \label{fig:appx_ghost_1290}
\end{figure*}

\newpage
\section{Uncertainty Visualisation}
\label{sec:appx_uncertainty}

This appendix visualises the per-voxel aleatoric uncertainty $\hat{s}_v$ predicted by the decoder head, complementing the quantitative ablation of the uncertainty-aware blending mechanism (\cref{subsec:blending,tab:app_ablation}).

\begin{figure*}[!htbp]
    \centering
    \includegraphics[width=\linewidth,trim=0mm 0mm 0mm 20mm, clip]{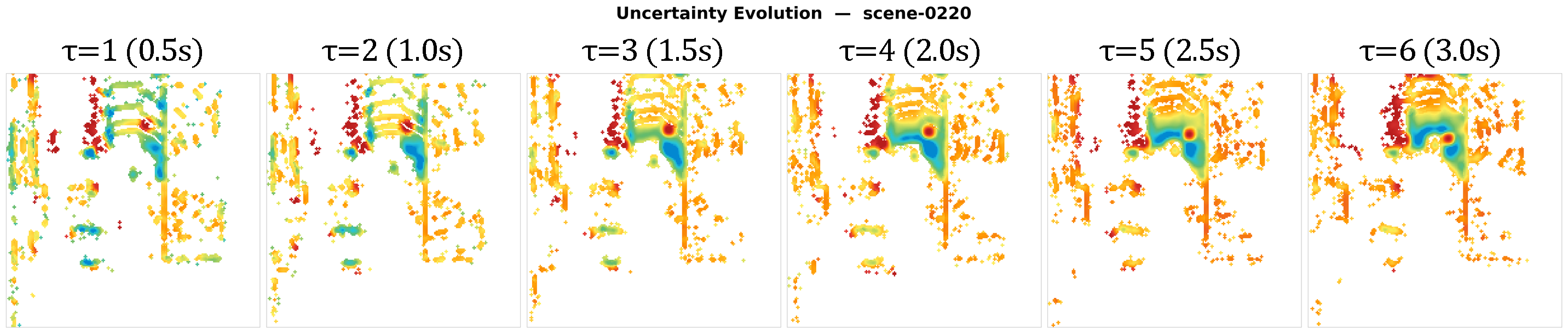}
    \caption{\textbf{BEV uncertainty evolution (nuScenes scene-0220, $\tau{=}1$ to $\tau{=}6$).}
    Each panel shows the BEV occupancy coloured by the predicted aleatoric uncertainty $\hat{s}_v$ at horizon $\tau$.
    Cool colours (cyan/green) indicate low uncertainty; warm colours (orange/red) indicate high uncertainty.
    Uncertainty increases with horizon and concentrates on dynamic objects and far-field voxels. Similar behaviour appears in \Cref{fig:appx_scatter_unc}.}
    \label{fig:appx_unc_strip}
\end{figure*}

\paragraph{BEV uncertainty evolution.}
\Cref{fig:appx_unc_strip} shows BEV projections of predicted occupancy coloured by~$\hat{s}_v$ for scene-0220 across all six prediction horizons.
Three trends are consistent across the sequence:
\begin{itemize}[nosep,leftmargin=1.5em]
    \item \textbf{Near-field confidence.} At $\tau{=}1$ (0.5\,s), uncertainty is concentrated on dynamic objects and occupancy boundaries, while large static structures remain comparatively confident.
    \item \textbf{Horizon-dependent growth.} As $\tau$ increases from 1 to 6 (0.5\,s to 3.0\,s), high-uncertainty regions expand and intensify. By $\tau{=}4$, a turning vehicle near the centre of scene-0220 dominates the uncertainty map, reflecting the increasing ambiguity of its future trajectory.
    \item \textbf{Far-field diffusion.} Sparse peripheral voxels become uncertain earlier than dense near-field regions, consistent with the weaker LiDAR support available at long range.
\end{itemize}
These patterns are consistent with the ensemble rule in \cref{subsec:blending}: short-horizon predictions rely primarily on the decoder, whereas long-horizon and dynamic regions receive greater weight from the ODE branch.

\paragraph{3D voxel-level analysis.}
\Cref{fig:appx_scatter_unc} renders the same uncertainty signal in 3D for scene-0274, which makes the vertical distribution easier to inspect.
At short horizons, the predicted structure is vertically coherent and densely populated; as $\tau$ increases, dropout and insertion artefacts become more visible, especially in thin or weakly observed structures.
The uncertainty map tracks this degradation closely: voxels that remain at $\tau{=}6$ are predominantly assigned high uncertainty, indicating that decoder confidence deteriorates in step with forecast difficulty.
This behaviour is precisely what the ensemble mechanism exploits when assigning more weight to the ODE path at longer horizons.

\newpage
\section{Cross-Domain Qualitative Results on SemanticKITTI}
\label{sec:appx_kitti}

The following figures visualise zero-shot transfer results on SemanticKITTI sequences, using models trained exclusively on nuScenes with no fine-tuning.

\begin{figure*}[!htbp]
    \centering
    \includegraphics[width=\linewidth,trim=0mm 0mm 0mm 40mm, clip]{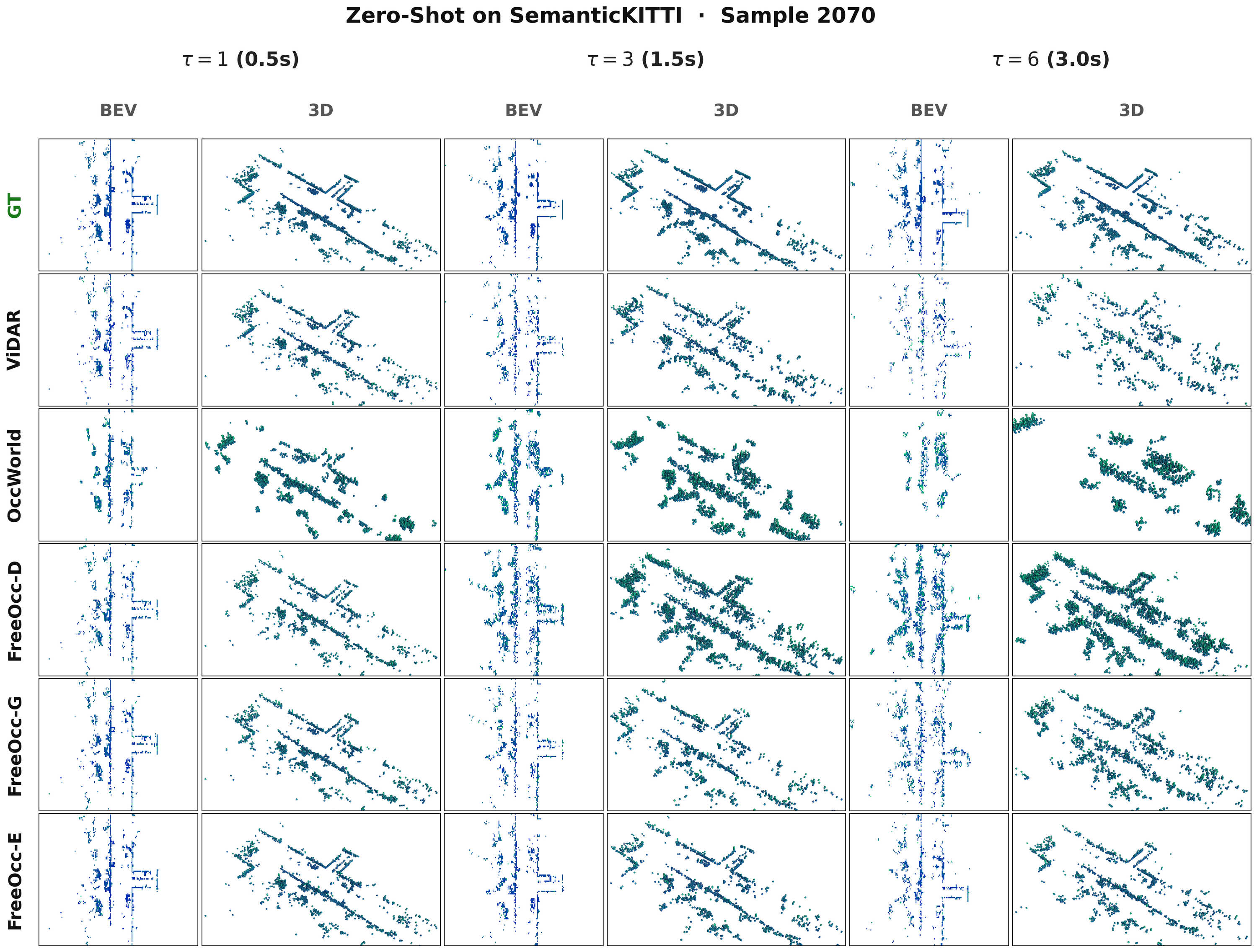}
    \caption{\textbf{Zero-shot SemanticKITTI predictions (sample 2070).}
    BEV and 3D occupancy predictions from a nuScenes-trained model evaluated on SemanticKITTI, comparing FreeOcc-E against OccWorld and ViDAR.
    Despite domain differences (64-beam LiDAR, different driving context, no fine-tuning), FreeOcc-E preserves road-level structure and major obstacles.
    The denser 64-beam input provides richer geometric detail than nuScenes's 32-beam data, and FreeOcc adapts without retraining, an advantage of operating on raw 3D geometry rather than appearance-dependent features.}
    \label{fig:appx_kitti_main}
\end{figure*}

\begin{figure*}[!htbp]
    \centering
    \includegraphics[width=\linewidth,trim=0mm 0mm 0mm 40mm, clip]{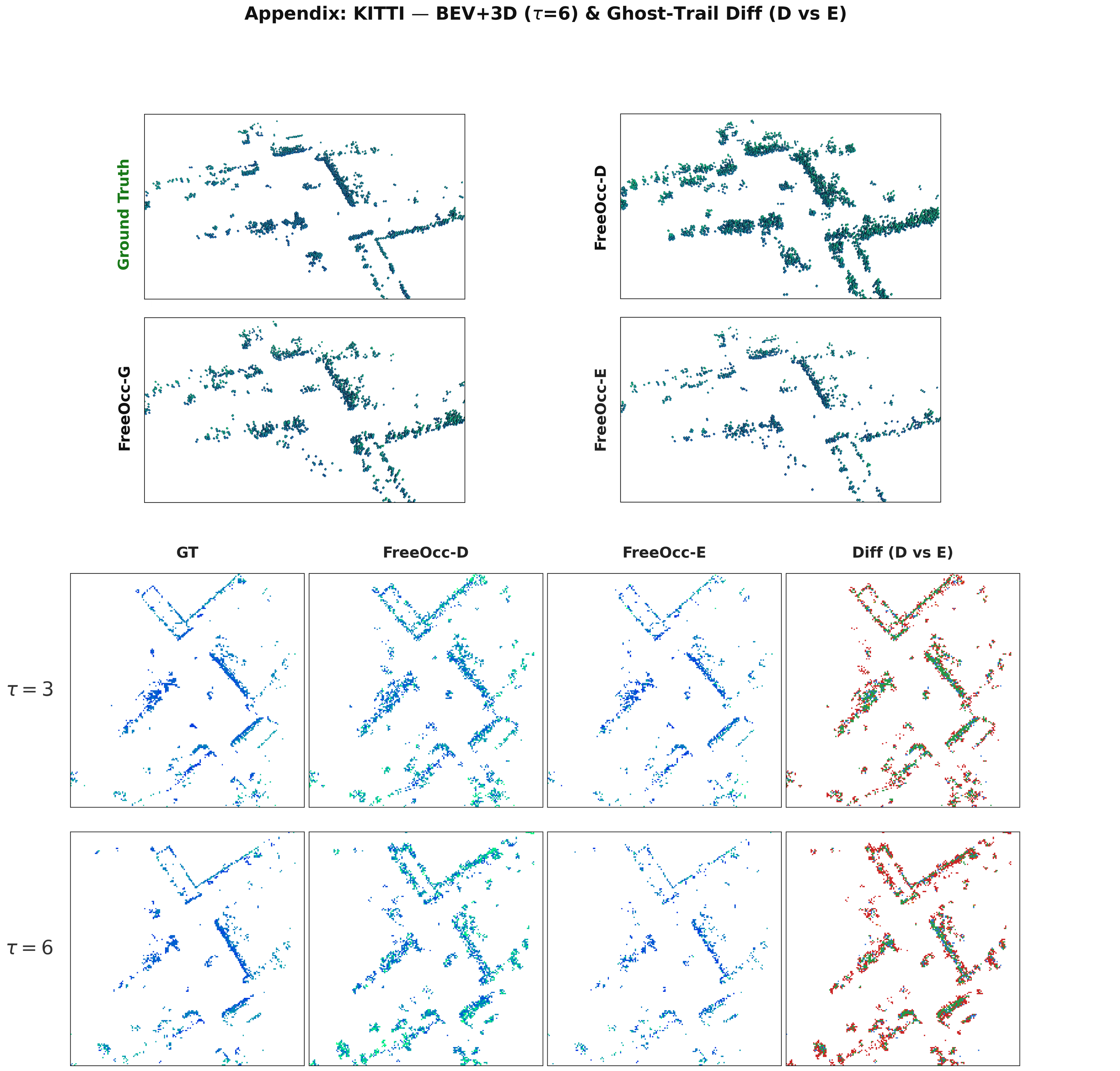}
    \caption{\textbf{Additional zero-shot SemanticKITTI scene (sample 1080) and noise suppression analysis.}
    Top: 3D occupancy predictions at the maximum forecasting horizon ($\tau=6$, 3.0s), comparing our model variants (FreeOcc-D, -G, -E) against Ground Truth.
    Bottom: BEV temporal evolution at $\tau=3$ and $\tau=6$. 
    Road geometry, vegetation boundaries, and building walls are consistently captured, reinforcing the cross-domain robustness reported in \cref{tab:zeroshot}.}
    \label{fig:appx_kitti_gallery}
\end{figure*}

\newpage
\section{Robustness Qualitative Results}
\label{sec:appx_robustness}

These visualizations illustrate the effect of degraded sensor conditions (timestamp jitter + 30\% frame dropout) on model predictions, corresponding to the quantitative analysis in \cref{tab:app_sensor_robustness}.

\begin{figure*}[!htbp]
    \centering
    \includegraphics[width=\linewidth,trim=0mm 0mm 0mm 40mm, clip]{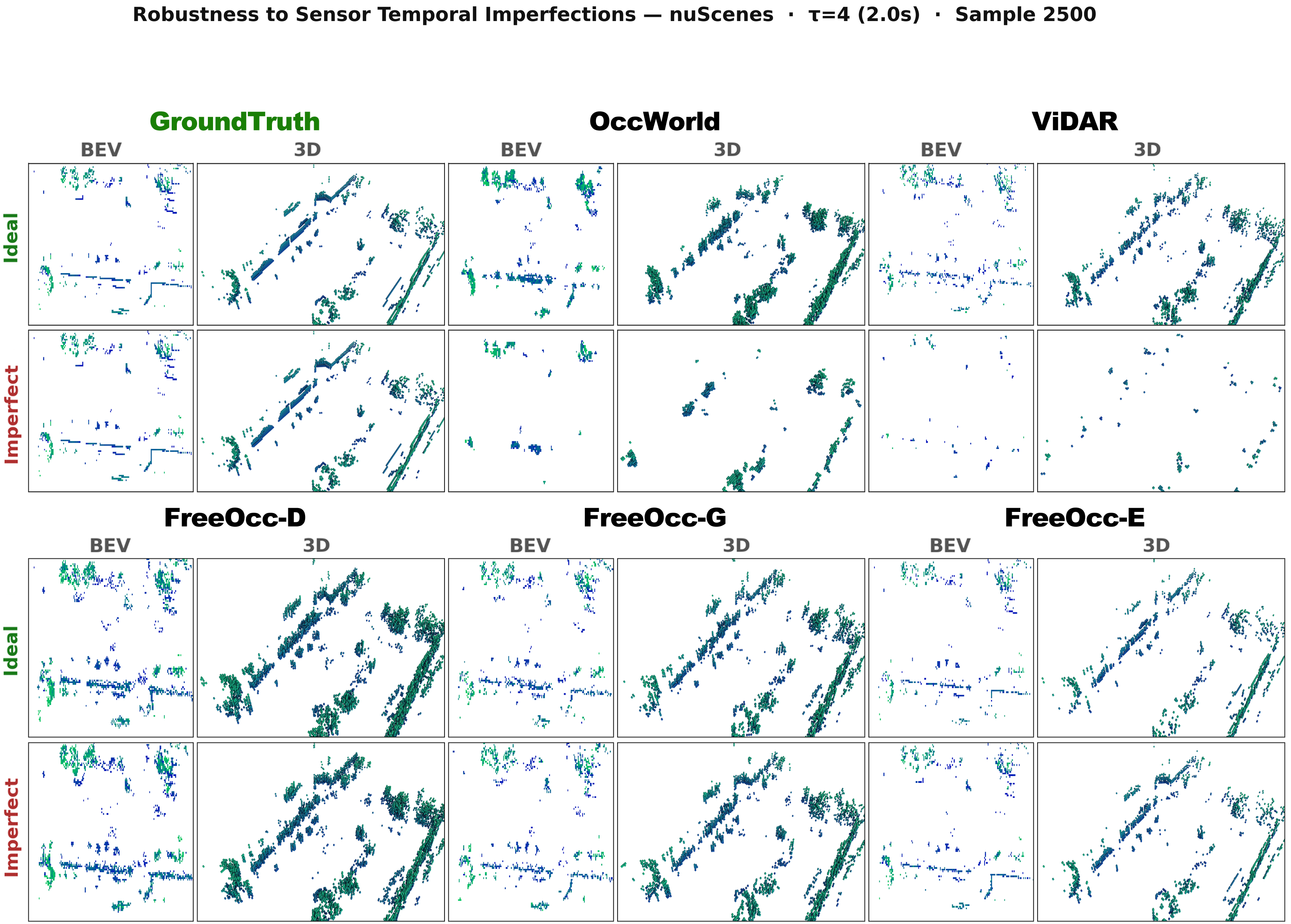}
    \caption{\textbf{Robustness on nuScenes ($\tau=4$, 2.0s, sample 2500).}
    Occupancy predictions under ideal and imperfect sensor conditions.
    FreeOcc-E retains its predictions under severe temporal perturbations, while OccWorld and ViDAR break into fragments and hallucinated clusters when temporal input is degraded.}
    \label{fig:appx_robust_nu_2500}
\end{figure*}

\begin{figure*}[!htbp]
    \centering
    \includegraphics[width=\linewidth,trim=0mm 0mm 0mm 30mm, clip]{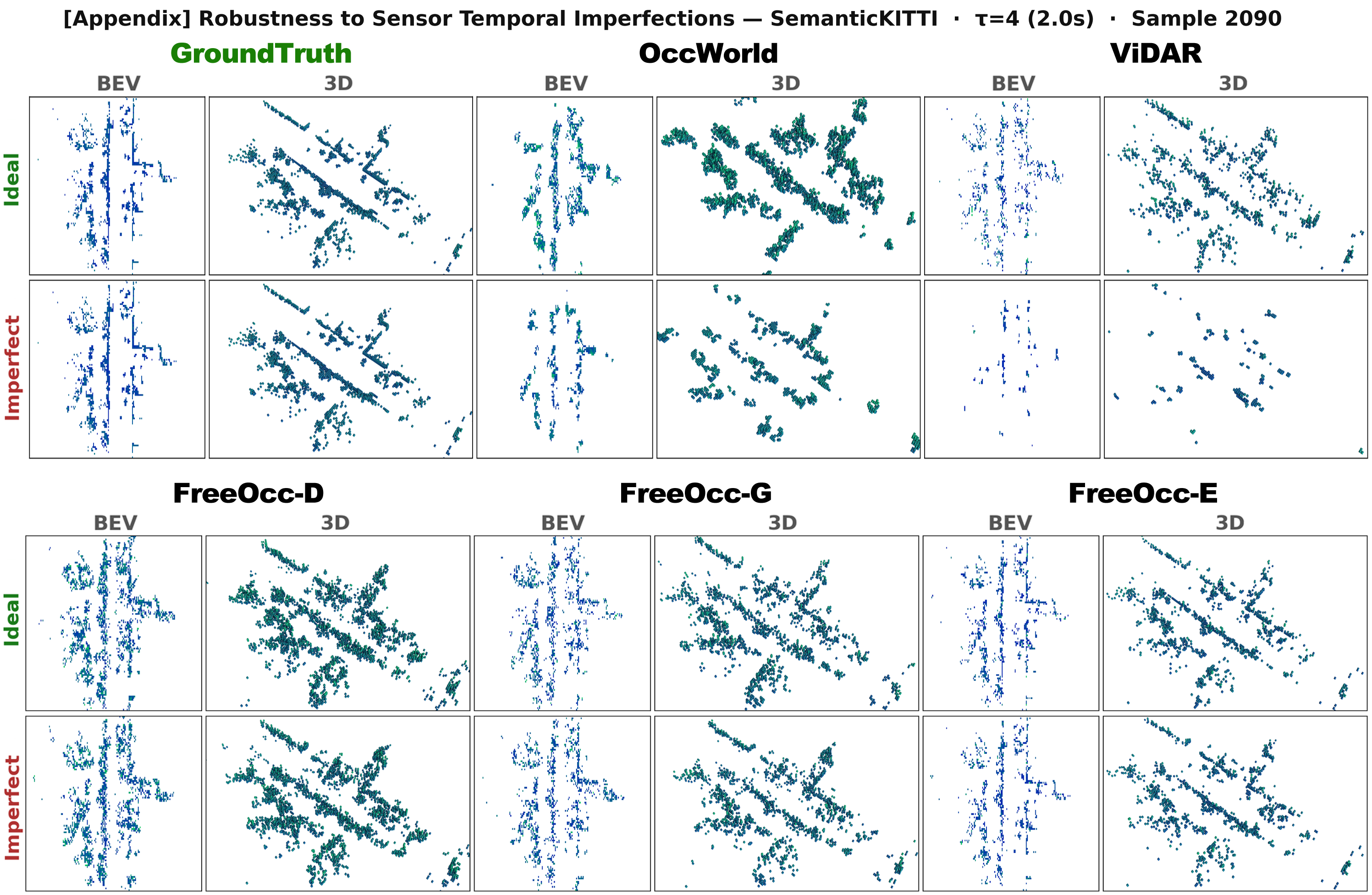}
    \caption{\textbf{Cross-domain robustness on SemanticKITTI ($\tau=4$, 2.0s, sample 2090).}
    Zero-shot predictions under ideal (top two rows) and imperfect (bottom two rows) sensor conditions.
    The cross-domain robustness gap compounds both the sensor-domain shift (nuScenes$\to$SemanticKITTI) and the temporal perturbation, yet FreeOcc-E retains recognisable road and object geometry in both cases, while OccWorld and ViDAR suffer severe structural collapse.}
    \label{fig:appx_robust_kt}
\end{figure*}

\begin{figure*}[!htbp]
    \centering
    \includegraphics[width=\linewidth,trim=0mm 0mm 0mm 30mm, clip]{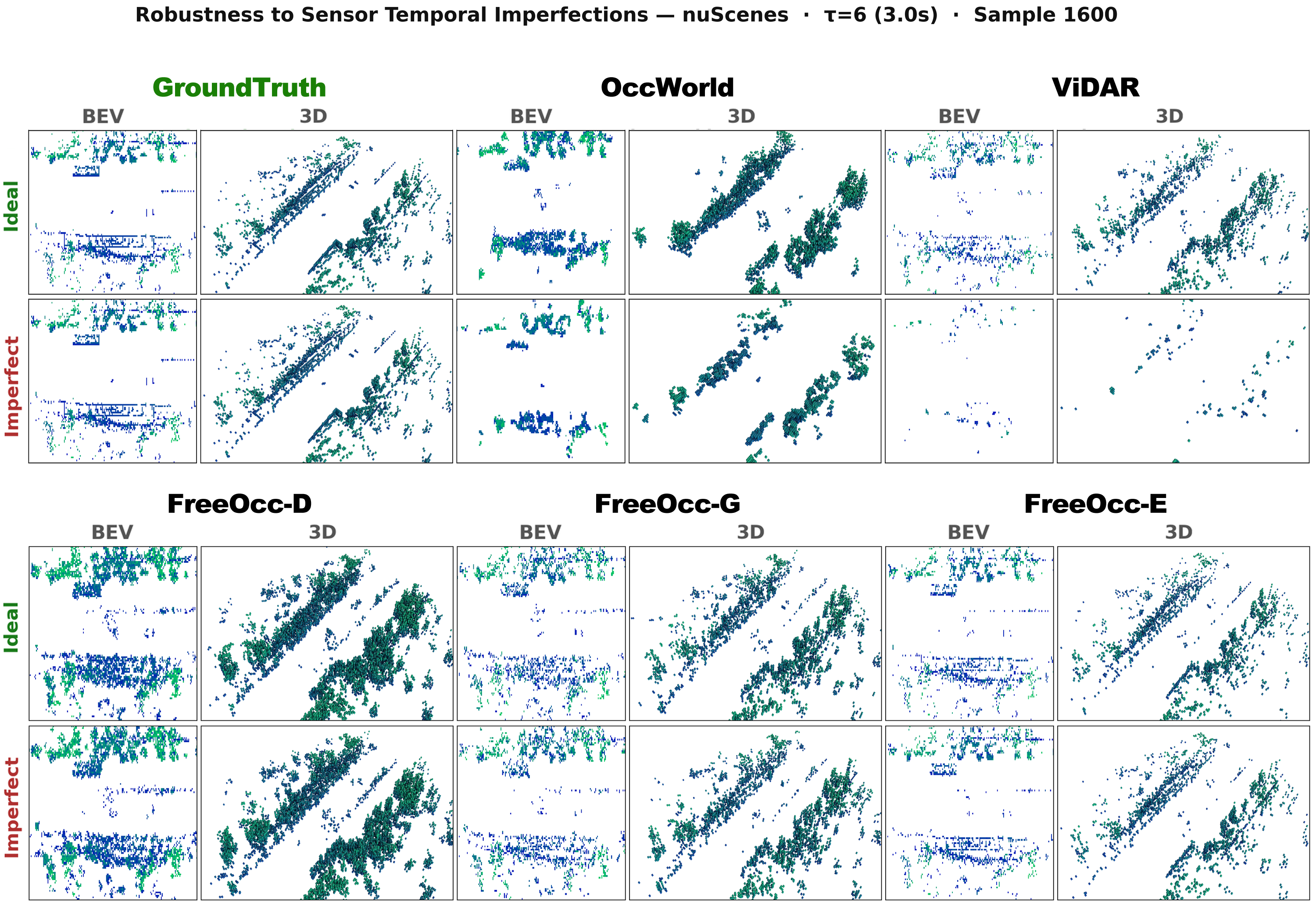}
    \caption{\textbf{Robustness on nuScenes at $\tau{=}6$ (3.0\,s, sample 1600).}
    Long-horizon predictions under ideal (top) and imperfect (bottom) sensor conditions.
    At the maximum forecasting horizon, OccWorld and ViDAR dissolve into discontiguous fragments, whereas FreeOcc-E preserves the road layout and principal object silhouettes.  The gap between ideal and imperfect rows is noticeably smaller for FreeOcc-E than for OccWorld or ViDAR, consistent with the quantitative retention rate in \cref{tab:app_sensor_robustness}.}
    \label{fig:appx_robust_nu_t6}
\end{figure*}

\begin{figure*}[!htbp]
    \centering
    \includegraphics[width=\linewidth,trim=0mm 0mm 0mm 30mm, clip]{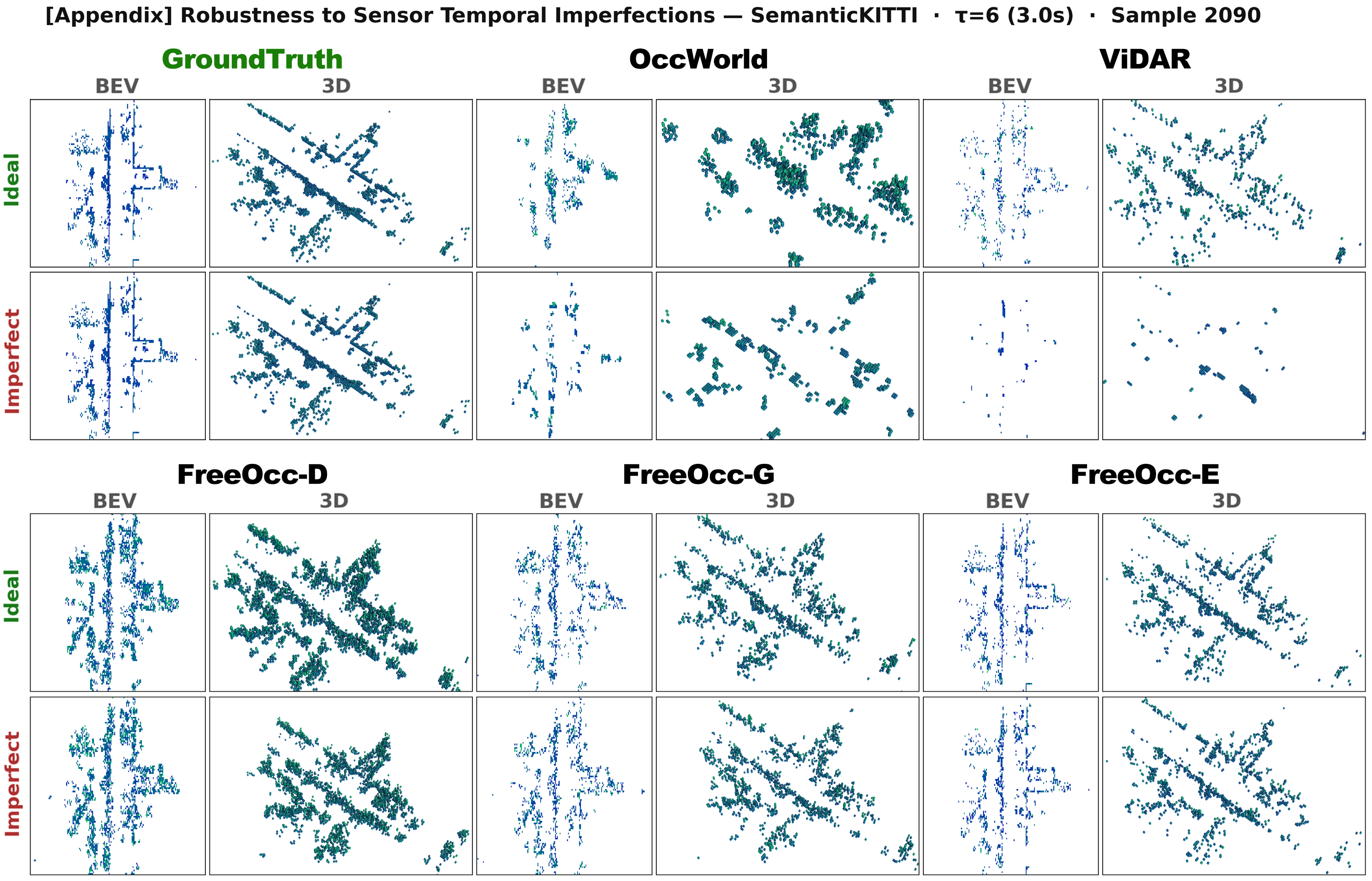}
    \caption{\textbf{Cross-domain robustness on SemanticKITTI at $\tau{=}6$ (3.0\,s, sample 2090).}
    The most challenging evaluation setting: simultaneous cross-domain transfer, long-horizon prediction, and degraded sensor input.
    FreeOcc-E still recovers the overall road layout and major object boundaries, whereas OccWorld and ViDAR suffer near-complete structural collapse.}
    \label{fig:appx_robust_kt_t6}
\end{figure*}

\newpage
\section{Latency Under Adverse Conditions}
\label{sec:appx_latency}

A practical question is whether latency remains stable on visually and geometrically difficult scenes, such as rain, reflections, and dense intersections. Here we evaluate runtime directly on a challenging nuScenes rain sequence and report the observed spread.

\begin{figure}[!htbp]
    \centering
    \includegraphics[width=\linewidth]{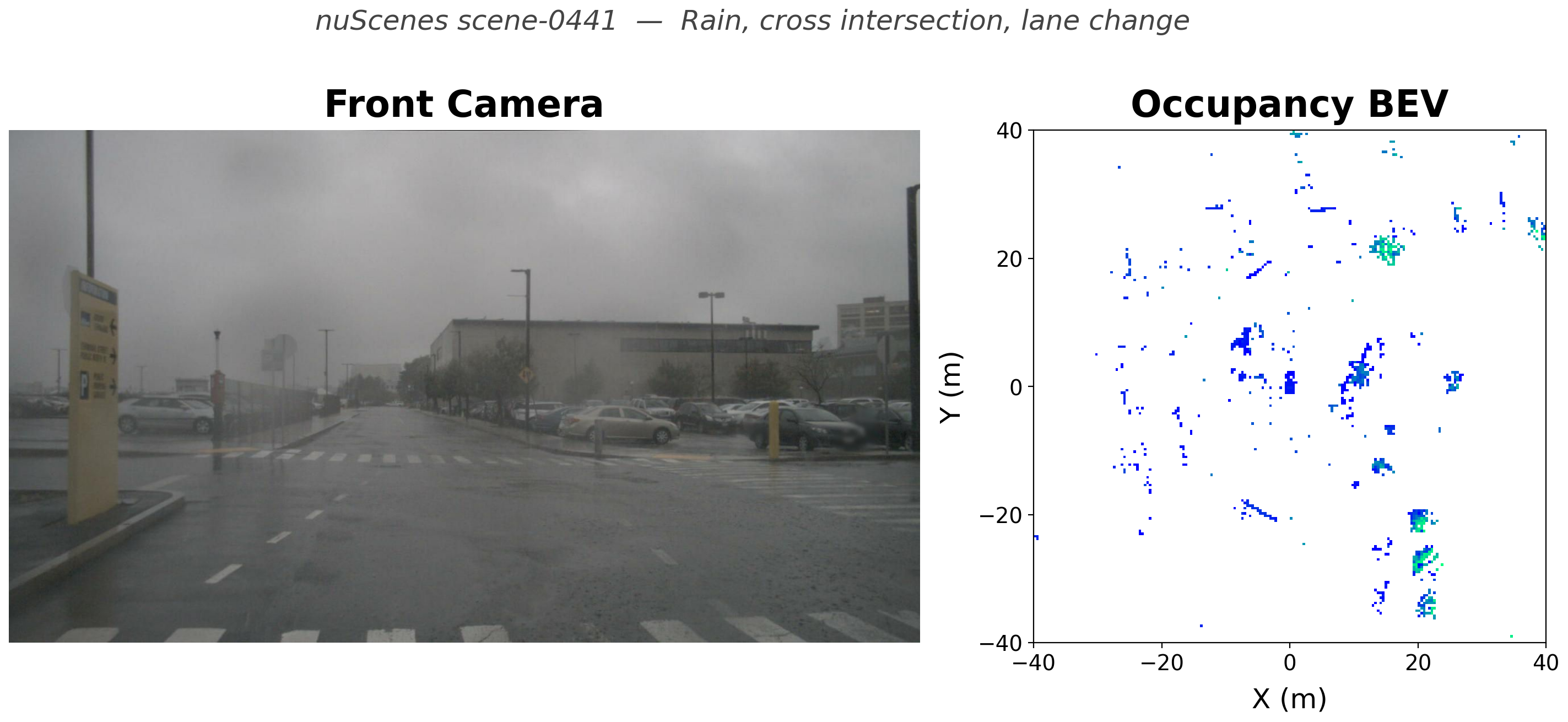}
    \caption{\textbf{Rain scene overview (nuScenes scene-0441).}  Left: front camera image showing wet road, reflections, and rain droplets.  Right: LiDAR BEV coloured by height; the intersection geometry and parked vehicles are clearly visible despite rain-induced noise in the point cloud.}
    \label{fig:rain_scene}
\end{figure}

\paragraph{Evaluation protocol.}
\Cref{fig:rain_scene} shows a representative keyframe from nuScenes \texttt{scene-0441} (rain, cross-intersection, lane change), one of the most demanding sequences in the dataset. We measure per-frame latency over all 40 keyframes and report the mean and worst-case runtime for FreeOcc-D and FreeOcc-E in \cref{tab:latency_rain}. OccWorld is included as an ideal-condition reference from our benchmark run.

\paragraph{Observed latency stability.}
FreeOcc-E maintains a mean latency of 37.3\,ms (26.8\,FPS), with worst-case latency both at 39.1\,ms. This spread is small and remains within roughly 2\,ms of the dry-weather average reported in the main paper. FreeOcc-D is similarly stable at 22.1\,ms (45.2\,FPS). Even in this adverse sequence, FreeOcc-E remains faster than our OccWorld reference measured under ideal conditions (56\,ms, 17.8\,FPS).

\paragraph{Interpretation.}

These results suggest that, in our implementation, adverse weather and dense traffic do not materially bottleneck runtime. In FreeOcc-E, the fast discriminative decoder first processes the scene and generates the uncertainty map, which then selectively routes only the highly ambiguous regions to the continuous ODE solver. The stability shown in \cref{tab:latency_rain} indicates that even in challenging conditions like rain, the decoder remains confident enough to resolve the vast majority of the environment (e.g., road layout and free space). Consequently, the additional computational overhead from routing more voxels to the ODE solver is tightly bounded, resulting in a worst-case latency penalty of only about 2 ms. We report these empirical statistics to demonstrate that our uncertainty-aware routing is robust and avoids unbounded computational spikes in unpredictable environments.

\begin{table}[!htbp]
    \centering
    \caption{\textbf{Inference latency on rain sequence} (nuScenes scene-0441, 40 keyframes).  Both FreeOcc variants maintain stable latency indistinguishable from dry-weather conditions.  FreeOcc-E's worst-case is still 1.4$\times$ faster than our OccWorld run under ideal conditions.}
    \label{tab:latency_rain}
    \small
    \begin{tabular}{l c c c c}
    \toprule
    Method & Mean (ms)  & Worst (ms) & FPS (mean) \\
    \midrule
    OccWorld (ideal, ref.) & 56.0 & --- & 17.8 \\
    \midrule
    FreeOcc-D (rain) & 22.1 & 23.8 & 45.2 \\
    FreeOcc-E (rain) & 37.3 & 39.1 & 26.8 \\
    \bottomrule
    \end{tabular}
\end{table}

\clearpage

\section{Trajectory Robustness Analysis}
\label{sec:appx_traj_robustness}

To verify that FreeOcc's advantage does not arise from privileged access to the exact future route, we conduct two analyses.

\paragraph{Sensitivity to trajectory error.}
We inject Gaussian noise into the future ego-poses, reporting the accumulated L2 position error at 3\,s.

\begin{table}[!htbp]
\centering
\caption{FreeOcc-E mIoU under injected future-trajectory error on nuScenes.}
\label{tab:traj_noise}
\small
\begin{tabular}{lcccccc}
\toprule
Traj.\ error @3\,s (L2) & 0.0\,m & 0.3\,m & 0.5\,m & 1.0\,m & 2.0\,m & 4.0\,m \\
\midrule
FreeOcc-E mIoU & 32.1 & 31.8 & 31.5 & 31.1 & 30.2 & 28.6 \\
\bottomrule
\end{tabular}
\end{table}

FreeOcc-E degrades gracefully. Across the typical planner-error range ($<$2\,m at 3\,s), it stays at 30.2 mIoU or above, well above OccWorld's 28.9.

\paragraph{Self-predicted trajectory.}
We replace the recorded route with a constant-velocity / constant-yaw-rate extrapolation from the recent history, using no future information.

\begin{table}[!htbp]
\centering
\caption{FreeOcc-E with a self-predicted trajectory (no future information).}
\label{tab:traj_selfpred}
\small
\begin{tabular}{lcc}
\toprule
Trajectory & mIoU & NFCD@3s \\
\midrule
Recorded route (paper setting) & 32.1 & 0.47 \\
Self-predicted from history & 31.2 & 0.51 \\
OccWorld (no external traj.) & 28.9 & --- \\
\bottomrule
\end{tabular}
\end{table}

Even without access to the future route, FreeOcc-E reaches 31.2 mIoU, still above OccWorld.

\section{Information-Matched Warp Baselines}
\label{sec:appx_warp_baselines}

To bound the contribution of the trajectory warp, we build two baselines on FreeOcc's own temporally aggregated present state (not on OccWorld). Warp+Fuse warps this state to the future ego frame and refines it with a light fusion head. Warp+Fuse+FOV additionally masks the warped result to the future sensor field of view. Neither uses the flow-matching branch or the uncertainty ensemble.

\begin{table}[!htbp]
\centering
\caption{Information-matched warp baselines versus FreeOcc on nuScenes.}
\label{tab:warp_baselines}
\small
\begin{tabular}{lccccc}
\toprule
Method & mIoU & Prec. & Rec. & F1 & NFCD@3s \\
\midrule
Static Copy & 15.9 & 34.0 & 23.5 & 27.8 & 1.11 \\
Warp+Fuse & 20.4 & 35.3 & 33.1 & 34.2 & 1.04 \\
Warp+Fuse+FOV & 20.8 & 36.5 & 33.4 & 34.9 & 0.98 \\
FreeOcc-D & 28.4 & 41.3 & 49.4 & 44.9 & 0.63 \\
FreeOcc-E & 32.1 & 54.6 & 45.5 & 49.6 & 0.47 \\
\bottomrule
\end{tabular}
\end{table}

Both warp baselines trail FreeOcc-E by at least 11.3 mIoU, attributing the gap to the flow-matching branch and the uncertainty ensemble rather than the trajectory warp.

\section{Static vs.\ Dynamic Breakdown}
\label{sec:appx_static_dynamic}

We define a ground-truth occupied voxel as \emph{dynamic} if it falls inside the BEV footprint of a nuScenes 3D box of a movable class (vehicle, human, cycle) moving faster than 0.5\,m/s, and \emph{static} otherwise. According to our statistics on the nuScenes dataset, dynamic voxels constitute only about 4\% of occupied voxels.

\begin{table}[!htbp]
\centering
\caption{Static and dynamic mIoU on nuScenes.}
\label{tab:static_dynamic}
\small
\begin{tabular}{lcccc}
\toprule
Method & Static mIoU & Dynamic mIoU & Overall & Dyn.\ margin vs OccWorld \\
\midrule
Warp+Fuse+FOV & 26.8 & 4.2 & 20.8 & $-10.3$ \\
OccWorld & 37.2 & 14.5 & 28.9 & --- \\
FreeOcc-E & 37.5 & 19.6 & 32.1 & $+5.1$ \\
\bottomrule
\end{tabular}
\end{table}

FreeOcc and OccWorld are essentially tied on static voxels (37.5 vs 37.2), so the trajectory warp gives no meaningful static advantage. The entire FreeOcc gain is on dynamic voxels ($+5.1$), where the ego-warp offers no help.

\section{Comparison with SparseWorld-TC}
\label{sec:appx_sparseworld}

SparseWorld-TC~\cite{du2025sparseworld} is a camera-based method designed for Occ3D semantic completion. We run the released model and evaluate it at the $\{1,2,3\}$\,s horizons common to both methods, using our sensor-grounded metrics. We also include FreeOcc-Cam, a camera variant using a Lift-Splat-Shoot (LSS) encoder~\cite{philion2020lift} with our forecasting modules unchanged.

\begin{table}[!htbp]
\centering
\caption{Comparison with SparseWorld-TC on our metrics ($\{1,2,3\}$\,s horizons).}
\label{tab:sparseworld}
\small
\begin{tabular}{lcccc}
\toprule
Method & Modality & mIoU & NFCD@3s & FPS \\
\midrule
SparseWorld-TC & Camera & 23.7 & 0.82 & 9.8 \\
FreeOcc-Cam (ours) & Camera & 25.6 & 0.69 & 19.4 \\
FreeOcc-E (ours) & LiDAR & 30.2 & 0.47 & 27.0 \\
\bottomrule
\end{tabular}
\end{table}

FreeOcc-E leads on all metrics. The same-modality FreeOcc-Cam also surpasses SparseWorld-TC while running about twice as fast.

\section{Camera Modality Ablation}
\label{sec:appx_camera_ablation}

To disentangle the LiDAR modality from the forecasting modules, we build FreeOcc-Cam by swapping only the LiDAR encoder for an LSS~\cite{philion2020lift} image-to-BEV encoder. All other components are unchanged. As a control, the camera baseline shares the same LSS encoder but replaces our forecasting stack with a plain feed-forward occupancy head (no ego-motion warp, temporal aggregation, flow matching, or ensemble). Both are trained on the same data.

\begin{table}[!htbp]
\centering
\caption{Isolating the modality from the forecasting modules (6-horizon average on nuScenes).}
\label{tab:camera_ablation}
\small
\begin{tabular}{lcc}
\toprule
Configuration & Modality & mIoU \\
\midrule
FreeOcc (full) & LiDAR & 32.1 \\
FreeOcc-Cam & Camera & 26.5 \\
Camera baseline (no warp/CFM/ensemble) & Camera & 21.9 \\
\bottomrule
\end{tabular}
\end{table}

LiDAR exceeds camera by 5.6 mIoU, and FreeOcc-Cam exceeds the camera baseline by 4.6 mIoU. The forecasting modules thus add value independent of modality.

\end{document}